\pdfoutput=1

\documentclass[11pt]{article}

\usepackage[final]{acl}
\usepackage{adjustbox}
\usepackage{times}
\usepackage{latexsym}

\usepackage[T1]{fontenc}

\usepackage[utf8]{inputenc}

\usepackage{microtype}

\usepackage{inconsolata}

\usepackage{graphicx}

\usepackage{amsfonts}
\usepackage{amsmath}
\usepackage{amsthm}
\usepackage{breqn}  
\usepackage{algorithm}
\usepackage{algpseudocode}
\usepackage{soul}
\usepackage{booktabs} 
\usepackage{multirow} 
\usepackage{rotating} 
\usepackage{adjustbox} 
\usepackage{cancel}

\usepackage{color, colortbl}

\newtheorem{theorem}{Theorem}
\newtheorem{lemma}{Lemma} 
\newtheorem{definition}{Definition} 

\definecolor{light-gray}{gray}{0.95} 

\title{LAVa: Layer-wise KV Cache Eviction with Dynamic Budget Allocation}

\author{Yiqun Shen\textsuperscript{1}\textsuperscript{2}\footnotemark[1]
\qquad
Song Yuan\textsuperscript{3}
\qquad
Zhengze Zhang\textsuperscript{1}\textsuperscript{2}
\qquad
Xiaoliang Wang\textsuperscript{1}\textsuperscript{2}
\\ \bf
Daxin Jiang\textsuperscript{3}
\qquad
Cam Tu Nguyen\textsuperscript{1}\textsuperscript{2}\footnotemark[2]  \\
         \textsuperscript{1}State Key Laboratory for Novel Software Technology, Nanjing University \\ \textsuperscript{2}School of Artificial Intelligence, Nanjing University \quad
         \textsuperscript{3}Stepfun\\
         \texttt{yiqunshen@smail.nju.edu.cn}\\
         \texttt{ncamtu@nju.edu.cn}\\
         }

\begin{document}
\maketitle
\renewcommand{\thefootnote}{\fnsymbol{footnote}}
\footnotetext[1]{This work was done during Shen's internship at Stepfun.}
\footnotetext[2]{Corresponding author}

\begin{abstract}
KV Cache is commonly used to accelerate LLM inference with long contexts, yet its high memory demand drives the need for cache compression. Existing compression methods, however, are largely heuristic and lack dynamic budget allocation. To address this limitation, we introduce a unified framework for cache compression by minimizing information loss in Transformer residual streams. Building on it, we analyze the layer attention output loss and derive a new metric to compare cache entries across heads, enabling layer-wise compression with dynamic head budgets. Additionally, by contrasting cross-layer information, we also achieve dynamic layer budgets. {\color{black}LAVa is the first unified strategy for cache eviction and dynamic budget allocation that, unlike prior methods, does not rely on training or the combination of multiple strategies.} Experiments with benchmarks (LongBench, Needle-In-A-Haystack, Ruler, and InfiniteBench) demonstrate its superiority. {\color{black}Moreover, our experiments reveal a new insight: dynamic layer budgets are crucial for generation tasks (e.g., code completion), while dynamic head budgets play a key role in extraction tasks (e.g., extractive QA).} As a fully dynamic compression method, LAVa consistently maintains top performance across task types. Our code is available at https://github.com/MGDDestiny/Lava.
\end{abstract}

\section{Introduction}
Large language models (LLMs) have shown remarkable capability in handling long-text scenarios, enabling advancements in tasks such as question answering \cite{kamalloo-etal-2023-evaluating}, code generation \cite{Guo2023LongCoderAL}, and multi-turn dialogues \cite{vicuna2023}. To further enhance external knowledge integration, state-of-the-art models like Claude 3.5 \cite{TheC3}, GPT-4 \cite{openai2024gpt4technicalreport}, and Qwen2.5 Max \cite{qwen2025qwen25technicalreport} have extended their context lengths beyond 128K tokens. However, supporting such long contexts comes with increased computational challenges. One common approach to accelerating LLM inference is caching Key and Value vectors (KV Cache), but its high memory demand necessitates efficient cache compression techniques.


While existing compression methods have shown promise, they are largely heuristic, relying on statistical measures such as accumulated attention scores \cite{zhang2023h2o, li2024snapkv}. These metrics are derived from empirical observations rather than a theoretical foundation. Additionally, although dynamic head allocation \cite{feng2024ada} and dynamic layer allocation \cite{qin2025cake} have been explored, no method, to our knowledge, fully adapts head and layer budgets.

\begin{figure*}
    \centering
    \includegraphics[width=0.90\linewidth]{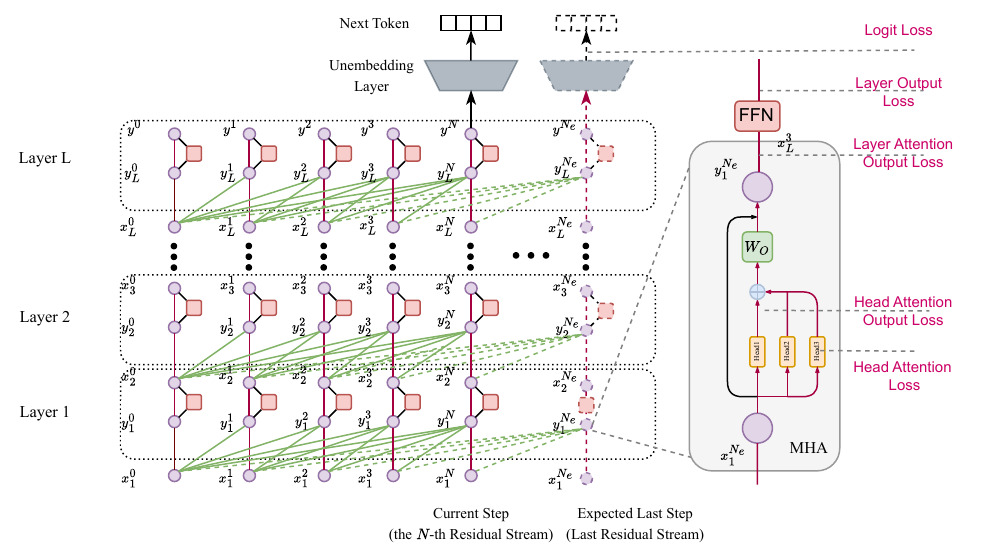}
    \caption{Information flow in decoder-only LLMs. The decoding process can be seen as operating on the current \textit{residual stream}. Each residual stream (red lines) corresponds to one token, and is considered as a \textit{communication channel}. Attention heads copy information from past residual streams to the current one (green lines) .}
    \label{fig:transformer-residual-streams}
\end{figure*}

To address this gap, we propose a unified framework for cache compression, which is formulated through the lens of minimizing information loss in Transformer residual streams (see Figure~\ref{fig:transformer-residual-streams}, and Sec. 3). Many existing methods can be formulated within our framework. Specifically, context compression methods \cite{qin2024dodo,qin2024nugget} aim to minimize global information loss at the logits layer. KV Cache compression methods \cite{zhang2023h2o,cai2024pyramidkv,qin2025cake} primarily focus on local information loss at the head or layer levels.

Our framework provides a principled approach to designing new algorithms. This paper introduces a novel method based on \textit{Layer Attention Output Loss}, which measures the impact of compression on the information retained in each layer after multi-head attention. The layer-wise loss function provides a balanced perspective on both local information within layers and global information flow across layers. Within each layer, the loss function guides the design of a scoring mechanism to assess token importance across heads, allowing for simultaneous head budget allocation and cache eviction. Across layers, it enables dynamic layer budget allocation by comparing information between layers. Our method is theoretically grounded, and significantly simpler than CAKE, the only training-free method with dynamic layer budgets.

Extensive experiments were conducted using various LLM series on the LongBench and Needle in a Haystack benchmarks. The results consistently demonstrate LAVa's strong ability to preserve the model's long-text comprehension under various memory constraints. Additionally, compared to a full cache implementation of FlashAttention-2, LAVa significantly reduces memory consumption while simultaneously reducing latency (9$\times$ faster decoding for 128K-token sequences). Our empirical findings highlight that dynamic layer budgets are essential for generation tasks, while dynamic head budgets are crucial for text extraction tasks. Achieving dynamic budget allocation at both the head and layer levels is key to optimizing performance across different tasks.

Our Contributions: 1) We introduce a \textbf{principled framework for KV Cache eviction} by analyzing the information flow through Transformer residual streams, accounting for information loss at various points during decoding. 2) Building on this framework and the notion of information loss at the layer-wise attention output, we propose LAVa—a unified method that simultaneously performs KV cache eviction and dynamic budget allocation. To the best of our knowledge, LAVa is the first training-free method to achieve dynamic budget allocation without relying on multiple combined metrics, making it simple for practical purposes. 3) Evaluations on LongBench, Needle in a Haystack, Ruler and InfiniteBench demonstrate that our simple method \textbf{outperforms strong baselines}. 4) Experiments reveal new insights into the role of dynamic budget allocation across different tasks, offering guidance for the adaptive selection of strategies.


\section{The Information Flow of LLM Decoding Process with KV Cache} 
\label{preliminary}


KV Cache is initialized at \textit{prefilling} stage, which basically computes the Key and Value for tokens in the initial prompts in the standard way \cite{vaswani2017attention}. In the following, we assume that there exists a KV Cache of $(N-1)$ previous tokens and demonstrate how decoding is performed at step-$N$.


\paragraph{Notations} The LLM has $L$ layers, each has $H$ heads. The model and head dimensions are $d$ and $d_h=d/H$; $K_l, V_l$ are the KV Cache for the $l$-th layer up to the current time step (the $N$-th token), which are of $[H, (N-1), d_h]$ sizes. The full notation Table~\ref{notation table} is in Appendix~\ref{ref:notation}. 

\paragraph{Decoding Process} According to \cite{ferrando-voita-2024-information}, LLM decoding can be viewed as operating on the current ($N$-th) \textit{residual stream}, as illustrated in Figure \ref{fig:transformer-residual-streams}. Specifically, suppose that $x^N_l$ is the current input for layer $l$, we first calculate the corresponding $Q^N_l, K^N_l, V^N_l$ as follows:
\begin{equation}
    Q^N_l = x^N_lW_l^Q; K^N_l = x^N_lW_l^K; V^N_l = x^N_lW_l^V \nonumber
\end{equation}
where $Q^N_l, K^N_l, V^N_l$ are of size ($H\times 1 \times d_h$), containing $H$ head-wise caches. 
The layer-wise KV Cache is then updated as follows:
\begin{equation}
    K_l = Cat[K_l,K^N_l], V_l = Cat[V_l,V^N_l] \nonumber
\end{equation}
where $K_l, V_l$ are tensors of size ($H\times N\times d_h$), and $Cat$ indicates the concatenation operation.
We then calculate the attention scores of step-$N$ for layer-$l$:
\begin{equation}
    A^N_l = Cat_{h\in[H]}\left(A^N_{l,h}\right)   \notag
\label{layer-wise attn}
\end{equation}
where $A^N_{l,h}=Softmax(\frac{Q^N_{l,h}{(K_{l,h})^T}}{\sqrt{d_h}})$. Here, {\textbf{\textit{$A^N_{l,h}[i]$ indicates how much the token at step-$N$ attends to the token-$i$ ($i\leq N$)}}. Layer-$l$ attention output is calculated as follows:
\begin{align}
    &y^N_l = Cat_{h\in[H]}(A_{l,h}^NV_{l,h} )W_l^O \in \mathbb{R}^{1\times d} \notag
\end{align}

The layer output $x^N_{l+1}$ is calculated as $x^N_{l+1}=y^N_l+FFN(y^N_l)$, which is then passed as the input the next layer $l+1$. In the last layer, we exploit an un-embedding layer ($W^M\in\mathbb{R}^{d\times |\mathcal{V}|}$) to get the probability vector $p^N$ for next token sampling.

\section{A Principled Framework for KV Cache Eviction based on Information Loss}
\label{framework}

Given the KV Cache, compression can be seen as masking entries in the $KV$ tensors so that the attention heads cannot copy masked information to the later residual streams. Formally, one can define the attention mask $\mathcal{I}_{l,h}$ for layer-$l$ and head-$h$:
\begin{equation}
    \mathcal{I}_{l,h}[i] =
    \begin{cases} 
    1 & \text{if } K_{l,h}[i] \text{ and } V_{l,h}[i] \text{ are retained} \\
    0 & \text{evict } K_{l,h}[i] \text{ and } V_{l,h}[i]
    \nonumber\end{cases}    
\end{equation}

The goal is to find a KV Cache eviction policy so that to minimize the information loss for the logits at the last layer ($p^N$) for all subsequent residual streams (from $N$ to $N_e$; see Figure \ref{fig:transformer-residual-streams}). Let $\mathcal{P}$ denote this logit loss, and $\mathbb{B}$ be the memory constraint. The unified problem for budget allocation and cache eviction can be defined as follows:
\begin{align}
    &\min_{\mathcal{I},\mathcal{B}} \mathcal{P}(\textbf{x}^{1\ldots N}_1,\mathcal{I},\mathcal{B}) 
        \label{loss objective}\\
    \text{st.} &\sum_{i\in[N]}\mathcal{I}_{l,h}[i]=\mathcal{B}_{l,h}; \nonumber \\
    &\sum_{h\in[H]}\mathcal{B}_{l,h}=\mathcal{B}_l; 
    \sum_{l\in[L]}\mathcal{B}_{l} = \mathbb{B} \nonumber\\ 
    &\mathcal{I}_{l,h}[k] = 1 \text{ },\forall l,h; \text{ and } \forall k \in [N-w,N]\nonumber
\end{align}

Here, \(\mathcal{B}_{l,h}\) represents the budget for layer-$l$ and head-$h$,  \(\mathcal{B}_l\) denotes the total budget for layer-$l$. The final constraint ensures that the most recent tokens within a window of size \(w\) are retained for all heads, aligning with the common practice in the literature.

\begin{table*}[t]
\centering
\small
\begin{tabular}{@{}l|c|c|p{5.5cm}|p{2cm}}
\toprule
\multirow{2}{*}{\textbf{Methods}} & \multicolumn{2}{|c|}{\textbf{Budgets}} & \multirow{2}{*}{\textbf{Scoring Function}} & \multirow{2}{*}{\textbf{Loss}} \\ \cmidrule{2-3}
& $\mathcal{B}_{l,h}$ & $\mathcal{B}_{l}$ & & \\\midrule
SnapKV \cite{li2024snapkv} & $\mathcal{B}_l/H$ & $\mathbb{B}/L$ & Recent attention scores & \multirow{5}{=}{Head Attention} \\
 & & &   $s_{l,h}[i]=\frac{1}{w}\sum_{j=N-w}^NA_{l,h}^j[i]$, $\forall i<N-w$ & \\ 
CAKE \cite{qin2025cake} & $\mathcal{B}_l/H$ & Dynamic & Recent attention scores + attention shifts & \\
& & & $s_{l,h}[i]= \gamma \text{VAR}_{j=N-w}^N([A_{l,h}^j[i]))$ & \\
& & & $+\frac{1}{w}\sum_{j=N-w}^N A_{l,h}^{j}[i] $, $\forall i<N-w$  & \\
\midrule
AdaKV \cite{feng2024ada} & Dynamic & Fixed & Recent attention scores (like SnapKV) & \multirow{3}{=}{Layer Attention Output}\\
\textbf{LAVa (Ours)} & Dynamic & Dynamic & Recent attention scores $\times$ value norm & \\
& & & $s_{l,h}[i]=\frac{max_{k}\|V_{l,h}[k]\|_1}{w}\sum_{j=N-w}^NA_{l,h}^j[i]$ &\\
\bottomrule
\end{tabular}
\caption{Summary of representative methods for KV Cache compression. {\color{black}LAVa is the only method to support dynamic head ($\mathcal{B}_{l,h}$) and layer ($B_l$) budgets}. For the full table and more comparison, please refer to Appendix~\ref{descriptions}.}.
\label{tab:summary}
\end{table*}

As computing the loss over future, unseen tokens is impractical. To address this, we approximate the loss by considering only residual streams up to the current step \(N\). Considering the current step-$N$, one can define \(\mathcal{P}\) as the cross-entropy loss between \(p^N\) and \(\hat{p}^N\), which is the logit obtained with the attention mask \cite{qin2024dodo}. Additionaly, since the search space for the mask matrix is combinatorial,  we instead search for a scoring function \(s\), where \(s_{l,h}[i]\) assigns an importance score to token \(i\) at layer \(l\) and head \(h\). This scoring function allows us to greedily choose the least important entries to be masked $\mathcal{I}=Select(s, \mathcal{B})$. All in all, we have the following (surrogate) optimization problem:
\begin{align}
    &\min_{\mathcal{B}, s\in\mathcal{F}} \mathcal{P}(\textbf{x}^{1\ldots N}_1,s,\mathbb{B}) 
        \label{loss objective1}
\end{align}
where \(\mathcal{F}\) denotes the space of all scoring functions. The scoring function can be parameterized by a network \(\phi\), which is then found through offline training. This is the common approach employed in context compression methods \cite{qin2024dodo,qin2024nugget}.  

The aforementioned approach to minimizing \textit{Global Logit Loss} can be impractical for online inference when the scoring function is computationally expensive. A more feasible alternative is to focus on local information and apply localized KV Cache eviction. For instance, \textit{Head Attention Loss} can be used for head-wise eviction, a strategy adopted by most existing methods \cite{zhang2023h2o,li2024snapkv,qin2025cake}. In this case, the scoring functions are lightweight, relying on simple statistical features, like head-wise attention weights. Table \ref{tab:summary} summarizes how existing methods can be formalized within our framework, with further details provided in Appendix~\ref{descriptions}.

\section{LAVa: Layer-wise Cache Eviction with Dynamic Budget Allocation}
\label{methodology}

\subsection{Layer Attention Output Loss and the Scoring Function} 
The aforementioned framework provides a principled approach to designing new algorithms for KV Cache eviction. This section demonstrates the design of our novel algorithm based on \textit{Layer Attention Output Loss} (see Figure \ref{fig:transformer-residual-streams}). Specifically, we show how our scoring function is designed based on analyzing the upper bound of the loss and how we can exploit the scoring function for layer-wise cache eviction with dynamic budget allocation. 


\begin{lemma} Based on the $L_p$ norm, the layer attention output loss due to the attention mask $\mathcal{I}$ is measured for layer-$l$ at the current ($N$-th) residual stream as follows:
    \begin{align}\small
&\mathcal{P}(\textbf{x}^{1\ldots N}_1,\mathcal{I},\mathcal{B})=\|y_l^N-\hat{y}_l^N\|_p \\
    &= \left\|Cat_{h}\left[\left(A^N_{l,h}-\frac{A^N_{l,h}\odot \mathcal{I}_{l,h}}{\|A^N_{l,h}\odot\mathcal{I}_{l,h}\|_1}\right)V_{l,h}\right]W_l^O\right\|_p \nonumber
    \label{output perturbation}
\end{align}
where $\odot$ indicates element-wise multiplication and $\hat{y}_l^N$ indicates the layer attention output obtained by masking the KV Cache with $\mathcal{I}$ (equivalently, after KV Cache eviction).
\label{thm:layer-output-loss}
\end{lemma}

We then develop a new upper bound for the $L_1$ norm and provide the result in Theorem~\ref{upper bound of output}. The proof of these are both provided in Appendix~\ref{proof of upper bound}. 

\begin{theorem}
    The $L_1$ norm of the layer attention output loss can be bounded by:
\begin{dmath}
    \|y^N_l-\hat{y}^N_l\|_1 \\
     \leq 2\hat{C}\sum_{h\in[H]}\sum_{i\in[N]} A^N_{l,h}[i]\bar{V}_{l,h}\left(1-\mathcal{I}_{l,h}[i]\right)
     \label{eqn:upper bound}
\end{dmath}
 where $\hat{C}=\|{{W_l^O}^T}\|_1$ is a constant independent of any head or token within layer-$l$;  $\bar{V}_{l,h} = max_{k\in[N]}\|V_{l,h}[k]\|_1$ is a head-dependent value. 
\label{upper bound of output}
\end{theorem}

Given a fixed budget $B_l$, we consider a greedy algorithm that iteratively evicts one cache entry at a time until the cache budget is met. We evict  the entries with the smallest scores, given by the scoring function \( s_{l,h}[i] = A^N_{l,h}[i] \bar{V}_{l,h} \) to minimize the upper bound. Notably, this function incorporates a head-dependent value \( \bar{V}_{l,h} \), which should not be ignored when comparing KV Cache entries across different heads. This is different from AdaKV \cite{feng2024ada}, which considers the layer attention output loss yet does not take into account the values. This also provides a theoretical justification for the introduction of values into the scoring, which has been exploited heuristically in VATP \cite{guo-etal-2024-attention}. 4It is noted that we derive our metric through a detailed reasoning process, independently from VATP. The process is key to understanding the approximations we introduce, which enable future improvements. Moreover, recognizing that the metric is inherently grounded in a layer-wise perspective enables the design of dynamic budget allocation strategies, as demonstrated below. Empirical comparison to VATP is given in Table~\ref{tab:vatp ab}.

The scoring function \( s_{l,h}[i] = A^N_{l,h}[i] \bar{V}_{l,h} \) described earlier is based solely on analyzing the current residual stream (the $N$-th decoding step). To improve the performance for KV Cache eviction, we can incorporate information from all past residual streams similarly to H2O \cite{zhang2023h2o}. However, doing so introduces more computational overhead. Inspired by SnapKV \cite{li2024snapkv}, we instead incorporate information from recent \( w \) residual streams, yielding a new scoring function.

\begin{definition} Layer-wise Attention and Value (LAVa) score for the token-$i$ at layer-$l$, head-$h$ is defined as follows:
\begin{equation}
    s_{l,h}[i]=\frac{max_{k\in[N]}\|V_{l,h}[k]\|_1}{w}\sum_{j=N-w}^NA_{l,h}^j[i] \label{eqn:LAVa-score}
\end{equation}
\end{definition}
 
 \begin{algorithm}[t]
    \caption{LayerEvict: Layer-wise KV Cache Eviction based on LAVa Score}
\begin{algorithmic}[1]  
\State \textbf{Input:} Budget $\mathcal{B}_l$, KV Cache $K_l,V_l$
\State \textbf{Output:} Compressed KV Cache $\hat{K}_l,\hat{V}_l$
\State $s_l=[\ ]$
\For{$h = 1$ to $H$}
    \State Calculate $s_{l,h}[i],\forall i \notin [N-w,N]$ based on  Eq.~\ref{eqn:LAVa-score}
    \State $s_l.extend(s_{l,h})$
\EndFor
\Function{Evict}{$\mathcal{B}_l,s_l,K_l,V_l$}
    \State $\mathcal{S}_{l}$ $\leftarrow$ $\mathcal{B}_l$ largest entries based on $s_l$ \label{lavaevict function}
    \State $\mathcal{I}_{l,h}[k]=0, \forall (h,k) \notin \mathcal{S}_l$
    \For{$h = 1$ to $H$}
        \State $\hat{K}_{l,h}=K_{l,h}\odot \mathcal{I}_{l,h}$
        \State $\hat{V}_{l,h}=V_{l,h}\odot \mathcal{I}_{l,h}$
    \EndFor
    \State \textbf{Return} $\hat{K}_l,\hat{V}_l$
\EndFunction
\State \textbf{Return} $\Call{Evict}{\mathcal{ B}_l,s_l,K_l,V_l}$
\end{algorithmic}
\label{algorithm1}
\end{algorithm}

Based on this scoring function, we develop the layer-wise KV Cache eviction as outlined in Algorithm \ref{algorithm1}. Notably, we only evict entries outside the recent window $[N-w,N]$, effectively retaining the most recent tokens as specified by the final constraint in the optimization problem (Eq.~\ref{loss objective}).

\paragraph{Dynamic Head Budget.}  Our eviction method operates across attention heads within layer-$l$. Specifically, we flatten the LAVa scores from all heads in the layer into a one-dimensional array $s_l$ (Algorithm~\ref{algorithm1}, lines 3–6). We then compare and rank $\mathcal{B}_l$ cache entries across all heads for layer-wise eviction, effectively obtaining dynamic head budget while performing eviction.

\subsection{Layer Budget Allocation}
\label{layer allocation}

Recently, CAKE \cite{qin2025cake} and PyramidKV \cite{cai2024pyramidkv} have demonstrated the potential of allocating different budgets across layers. PyramidKV, however, is suboptimal as it assigns a fixed allocation pattern regardless of the input. In contrast, CAKE is prompt-dependent allocation (dynamic) but combines different scores for cache eviction and budget allocation, which requires tuning three hyperparameters, hindering its practical application. Below, we describe our hyperparameter-free algorithm based on the LaVa score.


\begin{algorithm}[t]
    \caption{LAVa: Dynamic Budget Allocation and Cache Eviction based on LAVa Score}
\begin{algorithmic}[1]  
\State \textbf{Input:} Total Budget $\mathbb{B}$, KV Cache $K,V$ Number of Layers $L$ 
\State \textbf{Output:} Compressed KV Cache $\hat{K},\hat{V}$
\State $s=[\ ], e=[\ ], \hat{K}=K,\hat{V}=V$

\For{$l = 1$ to $L$}
    \State Calculate  $s_{l}$ based on Eq.~\ref{eqn:LAVa-score}
    \State Calculate $e_l$ based on Eq.~\ref{normalized entropy 1}, \ref{normalized entropy 2}
    \State $s.append(s_l)$
    \State $e.append(e_l)$ 
    \For{$\tilde{l} =1 $ to $l$}
        \State $\mathcal{B}_{\tilde{l}} = \frac{e_{\tilde{l}}}{\sum_l{e_l}} \mathbb{B}$
        \State $\hat{K}_{\tilde{l}}, \hat{V}_{\tilde{l}} = \Call{Evict}{\mathcal{B}_{\tilde{l}},s_{\tilde{l}},\hat{K}_{\tilde{l}},\hat{V}_{\tilde{l}}}$
    \EndFor
\EndFor
\State \textbf{Return} $\hat{K},\hat{V}$
\end{algorithmic}
\label{complete algorithm}
\end{algorithm}

Our key idea is that layers with greater uncertainty in \textit{determining which cache entry to evict} should be allocated a larger budget. Specifically, based on the LAVa score, \textit{the probability of evicting token-$k$ at layer-$l$ and head-$h$} is obtained by normalizing the LAVa scoring values:
\begin{align}
\hat{s}_{l,h}[i]= \frac{s_{l,h}[i]}{\sum_{k,h}s_{l,h}[k]}
\label{normalized entropy 1}
\end{align}
The uncertainty for layer-$l$ is then measured by the \textit{normalized entropy} as follows:
\begin{align}
e_l=\frac{-\sum_{h,i}(\hat{s}_{l,h}[i]\log \hat{s}_{l,h}[i])}{H\times N}
\label{normalized entropy 2}
\end{align}


With such a measure, we can first initialize all KV Cache through prefilling, followed by cache compression. Unfortunately, this approach results in a high memory peak after prefilling (and before compression). To address this, the common practice is that we perform prefilling and cache eviction layer by layer. For dynamic layer budget allocation, we draw inspiration from CAKE: after prefilling layer-$l$, the lower layers ($<l$)  are recompressed. As a result, a lower layer is compressed multiple times using the same LAVa scores, but the budget is adjusted, becoming smaller over time as the memory is shared with more layers being prefilled. The complete algorithm is outlined in Algorithm \ref{complete algorithm}. 



\subsection{LLMs with GQA} 
Group Query Attention (GQA) \cite{ainslie2023gqa} is the technique most modern LLMs adopt due to its balance between performance loss and memory efficiency. In GQA, the KV Cache is compressed by sharing a single KV Cache among all heads within a group. When applying LAVa scores to GQA, we take a conservative approach: the group-wise score for a token is determined as the maximum of its head-wise scores within the corresponding group. In other words, we tend to retain the entry as long as it is important for at least one head within the group.

\section{Experiments}
\begin{table*}[t]
\centering
\begin{adjustbox}{max width=\textwidth}
\begin{tabular}{lccccccccccccccccccccc|c}
\toprule
\multirow{2}{*}{} & \multicolumn{4}{c}{\textbf{Single-Doc. QA}} & \multicolumn{4}{c}{\textbf{Multi-Doc. QA}} & \multicolumn{4}{c}{\textbf{Summarization}} & \multicolumn{4}{c}{\textbf{Few-shot Learning}} & \multicolumn{3}{c}{\textbf{Synthetic}} & \multicolumn{2}{c}{\textbf{Code}} & \multirow{1}{*}{}\\
\cmidrule(lr){2-5} \cmidrule(lr){6-9} \cmidrule(lr){10-13} \cmidrule(lr){14-17} \cmidrule(lr){18-20} \cmidrule(lr){21-22} 
& \rotatebox{-60}{NrtvQA} & \rotatebox{-60}{Qasper} & \rotatebox{-60}{MF-en} & \rotatebox{-60}{MF-zh} & \rotatebox{-60}{HotpotQA} & \rotatebox{-60}{2WikiMQA} & \rotatebox{-60}{Musique} & \rotatebox{-60}{Dureader} & \rotatebox{-60}{GovReport} & \rotatebox{-60}{QMSum} & \rotatebox{-60}{VCSUM} & \rotatebox{-60}{MultiNews} & \rotatebox{-60}{TREC} & \rotatebox{-60}{TriviaQA} & \rotatebox{-60}{SAMSum} & \rotatebox{-60}{LSHT} &  \rotatebox{-60}{PCount} & \rotatebox{-60}{PR-en} & \rotatebox{-60}{PR-zh} & \rotatebox{-60}{Lcc} & \rotatebox{-60}{RepoBench-P} & \rotatebox{-60}{Avg}\\
\midrule
\textbf{Full Cache} & 26.77 & 32.34 & 49.63 & 48.42 & 43.43 & 27.89 & 18.61 & 30.85 & 32.92 & 24.54 & 15.04 & 27.20 & 71.00 & 86.23 & 43.41 & 39.00 & 2.81 & 86.56 & 89.75 & 55.29 & 52.55 & 45.07\\
\midrule
\multicolumn{23}{c}{\textbf{$\mathbb{B}=128\mathrm{HL}$}}\\
\textbf{PyramidKV} & 20.01 & 19.23 & 43.81 & 32.37 & 35.62 & 22.34 & 14.38 & 17.53 & 18.95 & 21.91 & 11.07 & 20.87 & 47.00 & \textbf{85.34} & \textbf{40.21} & 19.25 & 2.86 & 65.60 & 59.49 & 49.52 & 45.67 & 34.51\\
\textbf{SnapKV} & \ul{20.99} & 19.65 & \textbf{45.04} & 32.02 & 36.48 & 22.19 & 14.04 & 17.68 & 18.83 & 21.36 & 10.91 & 20.29 & 45.00 & 84.10 & 40.01 & 19.75 & 3.06 & 64.48 & 60.50 & 49.84 & 45.27 & 34.42\\
\textbf{Ada-PyramidKV} & 20.21 & \ul{20.80} & 43.82 & 33.65 & \ul{37.21} & 22.99 & 14.93 & \ul{18.06} & \textbf{19.41} & 22.02 & 11.16 & 20.97 & \ul{52.00} & 83.93 & 39.97 & 20.00 & 2.81 & \textbf{72.73} & \ul{72.89} & 51.00 & 46.62 & \ul{36.22}\\
\textbf{Ada-SnapKV} & 20.61 & 20.56 & 44.03 & \textbf{34.03} & 36.39 & \ul{23.66} & \textbf{16.15} & 17.82 & 19.21 & 21.73 & \ul{11.25} & 20.35 & 50.00 & 84.32 & 39.82 & 19.75 & 3.87 & 69.11 & 70.52 & 50.21 & 46.85 & 35.82\\
\textbf{CAKE} & \textbf{21.01} & 20.16 & 44.08 & 32.52 & 36.16 & \textbf{23.89} & \ul{15.32} & 17.67 & 18.82 & \textbf{22.62} & 10.93 & \ul{21.03} & 47.00 & 85.14 & 39.90 & \ul{21.25} & 3.02 & 63.65 & 65.96 & \ul{51.81} & \ul{48.53} & 35.06\\
\textbf{LAVa (Ours)} & 19.57 & \textbf{21.11} & \ul{44.29} & \ul{33.91} & \textbf{38.29} & 23.59 & \ul{15.32} & \textbf{18.56} & \ul{19.33} & \ul{22.32} & \textbf{11.42} & \textbf{21.07} & \textbf{53.50} & \ul{85.20} & \ul{40.16} & \textbf{21.75} & 2.88 & \ul{69.87} & \textbf{74.75} & \textbf{51.94} & \textbf{48.92} & \textbf{36.74} \\
\midrule
\multicolumn{23}{c}{\textbf{$\mathbb{B}=256\mathrm{HL}$}}\\
\textbf{PyramidKV} & 20.79 & 22.74 & 45.90 & 35.72 & 38.63 & 24.02 & 15.97 & 18.99 & 21.61 & 22.34 & 11.02 & 22.24 & 58.00 & 84.06 & 40.52 & 22.75 & 2.96 & 74.70 & 83.83 & 51.85 & 48.86 & 38.23\\
\textbf{SnapKV} & 21.39 & 22.15 & 46.50 & 34.77 & \textbf{39.68} & \ul{25.01} & 14.86 & 19.11 & 21.61 & \ul{23.04} & 11.46 & 22.67 & 57.00 & 85.04 & 40.81 & 23.25 & 3.18 & 76.49 & 83.60 & 51.99 & 49.42 & 38.49\\
\textbf{Ada-PyramidKV} & \ul{22.61} & \ul{23.84} & \ul{47.65} & 36.56 & \ul{39.33} & 24.86 & \textbf{17.22} & \ul{19.65} & 21.22 & 22.54 & \ul{11.82} & 22.29 & \ul{64.00} & 84.93 & 40.36 & 24.50 & 3.40 & 77.39 & \ul{85.83} & 52.48 & 49.43 & \ul{39.43}\\
\textbf{Ada-SnapKV} & 21.63 & 23.55 & 47.51 & \ul{37.42} & 38.89 & 23.65 & 16.06 & 19.34 & \textbf{21.98} & \textbf{23.21} & 11.49 & 22.39 & \ul{64.00} & \textbf{86.33} & 40.54 & \ul{25.25} & 2.23 & \textbf{77.44} & 85.42 & 52.31 & 49.62 & 39.40\\
\textbf{CAKE} & 21.37 & 23.40 & 46.84 & 35.02 & 38.10 & 24.50 & 14.81 & 19.40 & 21.59 & 22.77 & 11.32 & \ul{22.68} & 55.00 & \ul{85.46} & \textbf{41.92} & 24.75 & 2.96 & 75.66 & \textbf{86.46} & \textbf{54.29} & \ul{51.38} & 38.84\\
\textbf{LAVa (Ours)} & \textbf{22.70} & \textbf{24.67} & \textbf{48.62} & \textbf{37.81} & \textbf{39.68} & \textbf{25.96} & \ul{16.77} & \textbf{20.26} & \ul{21.92} & 22.48 & \textbf{11.88} & \textbf{22.91} & \textbf{65.00} & 85.24 & \ul{41.28} & \textbf{26.75} & 2.88 & \ul{76.76} & 85.75 & \ul{54.17} & \textbf{51.77} &\textbf{40.12}\\
\midrule
\multicolumn{23}{c}{\textbf{$\mathbb{B}=512\mathrm{HL}$}}\\
\textbf{PyramidKV} & 23.57 & 24.84 & 48.74 & 39.54 & 38.90 & 25.22 & 17.40 & 20.42 & 23.04 & 23.24 & 11.91 & 24.19 & 66.50 & 86.07 & 41.06 & 28.00 & 3.29 & 87.29 & 88.83 & 53.77 & 50.42 & 41.15\\
\textbf{SnapKV} & 23.67 & \textbf{28.08} & \ul{49.40} & 40.25 & \ul{40.14} & 25.58 & 16.97 & 20.49 & \textbf{23.75} & \textbf{23.69} & 12.03 & 24.31 & 65.00 & 86.29 & 41.98 & 28.50 & 3.22 & 85.79 & 88.67 & 53.99 & 51.02 & 41.48\\
\textbf{Ada-PyramidKV} & 24.37 & 27.30 & 48.01 & 40.88 & 39.75 & 25.96 & \textbf{18.58} & 20.90 & 23.59 & 23.33 & 12.07 & 24.04 & \ul{67.50} & \textbf{86.44} & \textbf{42.58} & 31.50 & 3.38 & 85.88 & \ul{89.67} & 54.15 & 51.30 & 41.89\\
\textbf{Ada-SnapKV} & \ul{24.63} & 27.48 & 48.90 & \ul{41.28} & 39.84 & \ul{26.33} & 18.26 & \ul{20.91} & 23.59 & 23.51 & \ul{12.27} & 24.32 & \ul{67.50} & \ul{86.38} & 42.34 & \ul{32.50} & 2.98 & \textbf{87.65} & 89.17 & 54.39 & 51.03 & \ul{42.11}\\
\textbf{CAKE} & 22.76 & 27.54 & \textbf{49.47} & 41.27 & 38.17 & 25.85 & 17.26 & 20.60 & \ul{23.72} & \ul{23.65} & 11.95 & \ul{24.50} & 66.00 & 86.01 & \ul{42.56} & 29.50 & 3.45 & 86.79 & 88.75 & \textbf{56.40} & \ul{52.37} & 41.76\\
\textbf{LAVa (Ours)} & \textbf{25.01} & \ul{27.84} & 48.97 & \textbf{42.14} & \textbf{40.95} & \textbf{26.88} & \ul{18.33} & \textbf{21.12} & 23.59 & 23.59 & \textbf{12.28} & \textbf{24.51} & \textbf{68.50} & 86.34 & \ul{42.48} & \textbf{33.50}  & 2.90 & \ul{87.23} & \textbf{89.83} & \ul{55.83} & \textbf{52.85} & \textbf{42.59} \\
\midrule
\multicolumn{23}{c}{\textbf{$\mathbb{B}=1024\mathrm{HL}$}}\\
\textbf{PyramidKV} & \textbf{25.62} & 28.96 & 48.35 & 42.18 & 40.89 & 26.65 & \ul{19.69} & 21.96 & 25.10 & 23.57 & 12.58 & 25.42 & 68.50 & \textbf{86.30} & 41.92 & 35.50 & 2.98 & 86.77 & \ul{89.50} & 55.26 & 51.03 & 42.79\\
\textbf{SnapKV} & 24.80 & 30.17 & \ul{49.13} & \ul{43.23} & \ul{41.16} & 26.92 & 17.89 & 22.58 & 25.75 & 23.64 & 12.88 & 25.85 & 67.50 & \ul{86.25} & 42.56 & 36.00 & 2.88 & 88.10 & 88.92 & 55.23 & 51.38 & 43.00\\
\textbf{Ada-PyramidKV} & 24.98 & 29.92 & 47.97 & 41.43 & 40.83 & 26.98 & 19.42 & 22.45 & 25.46 & 23.58 & 12.94 & 25.61 & 68.50 & \textbf{86.30} & \textbf{42.84} & 35.50 & 2.89 & 88.18 & 89.25 & 54.51 & 51.32 & 42.90\\
\textbf{Ada-SnapKV} & 24.84 & 29.99 & \textbf{49.21} & 42.55 & 41.00 & \textbf{27.39} & 19.23 & \ul{23.23} & \ul{25.89} & \textbf{24.18} & 13.13 & 25.85 & \ul{69.00} & 86.23 & \textbf{42.84} & \ul{36.25} & 2.90 & \textbf{89.02} & \textbf{89.75} & 55.38 & 51.93 & 43.34\\
\textbf{CAKE} & 25.15 & \ul{30.34} & 49.00 & 43.08 & 40.86 & 26.70 & \textbf{19.93} & 23.07 & 25.82 & 23.72 & \ul{13.16} & \textbf{26.05} & 68.00 & \ul{86.25} & \ul{42.70} & 36.00 & 2.91 & \ul{88.60} & 88.75 & \ul{56.75} & \ul{53.26} & \ul{43.36}\\
\textbf{LAVa (Ours)} & \ul{25.59} & \textbf{31.21} & 48.27 & \textbf{43.43} & \textbf{41.92} & \ul{27.38} & 19.48 & \textbf{23.48} & \textbf{26.06} & \ul{23.86} & \textbf{13.38} & \ul{26.00} & \textbf{70.00} & 86.22 & 42.43 & \textbf{38.00} & 2.73 & 87.01 & 88.75 & \textbf{57.31} & \textbf{53.28} & \textbf{43.65} \\
\bottomrule
\end{tabular}
\end{adjustbox}
\caption{Final comparison based on Mistral-7B-Instruct-v0.2 among 21 datasets of LongBench. (Note: The best result is highlighted in \textbf{bold}, and the second is in \ul{underline}. Due to the negligible numerical values obtained from the passage count dataset, its results were excluded from the computation of the average scores.)}
\label{tab:mistral-longbench}
\end{table*}

\label{evaluation}
\subsection{Experimental Settings}
\label{settings}

\paragraph{Backbone LLMs.}
We evaluate three series of LLMs: Mistral-7B-Instruct-v0.2 \cite{jiang2023mistral7b}, Qwen2.5-7/14/32B-Instruct \cite{qwen2025qwen25technicalreport}, all with a context length of 32k and Llama3-8B-Instruct with 8k context length. These models are widely adopted for their moderate parameter sizes and strong performance all utilizing GQA \cite{ainslie2023gqa}.


\paragraph{Evaluation Benchmarks.}
To validate the effectiveness of our algorithm, we perform evaluation   \textit{\textbf{LongBench}} \cite{bai-etal-2024-longbench}, a bilingual, multi-task benchmark for long-context understanding. It comprises 21 datasets across six task categories in both English and Chinese, with an average length of 6,711 words (English) and 13,386 characters (Chinese). LongBench covers key long-text application areas, including single-document QA, multi-document QA, summarization, few-shot learning, synthetic tasks, and code completion. We also conduct experiments on \textit{\textbf{Needle In A Haystack}}~\cite{cai2024pyramidkv,liu-etal-2024-lost,needle2}, \textit{\textbf{Ruler}}~\cite{hsieh2024ruler} and \textit{\textbf{InfiniteBench}}~\cite{zhang-etal-2024-bench}, of which the results are given in Appendix~\ref{details of experiments}. 

\paragraph{Baseline Methods.}
We compare our methods against several baselines: PyramidKV, SnapKV, Ada-SnapKV, Ada-PyramidKV, and CAKE. Among these, PyramidKV and CAKE allow different layer budgets. AdaKV is derived from the layer attention output loss but relies solely on attention for its scoring function and does not incorporate dynamic layer budget allocation. Ada-SnapKV employs the same scoring function and uniform layer allocation as SnapKV but allows dynamic head budgets. Ada-PyramidKV follows the same approach but assigns fixed, varying budgets across layers like PyramidKV. 

Pooling operators, such as max pooling or average pooling, can be applied to token score vectors to smooth score variations across adjacent tokens \cite{li2024snapkv,cai2024pyramidkv,qin2025cake}. This strategy is also employed in the implementation of LAVa and all the baselines. For pooling operation, for all methods, we adopt maxpool function and set kernel size as 7. More information is given in Appendix~\ref{descriptions}, and for implementation details, please refer to Appendix~\ref{ref:implementation}.}



\subsection{Main Results}
\label{final evaluation}

Table \ref{tab:mistral-longbench} presents the results of Mistral-7B with different eviction policies on LongBench, revealing several key observations. First, \textit{LAVa outperforms all baselines across different budgets}, with a more pronounced advantage at smaller budgets. Second, among methods requiring no hyperparameter tuning (SnapKV, Ada-SnapKV, and LAVa), LAVa achieves the best performance, significantly surpassing others. For instance, at $\mathbb{B}=128\mathrm{HL}$, LAVa achieves an average score of 36.74, compared to Ada-SnapKV’s 35.82. And finally, \textit{LAVa and CAKE excel in code-related tasks}. On RepoBench-P with a $128\mathrm{HL}$ budget, LAVa (48.92) and CAKE (48.53) outperform Ada-SnapKV (46.85) by a significant margin. This is interesting given that Ada-SnapKV surpasses CAKE on average over 20 datasets. Similar trends are observed with the Qwen series and presented in Appendix~\ref{ref:results of our others longbench}. 

\begin{figure}[t]
    \centering
    \hspace{-0.2cm}\includegraphics[width=0.95\linewidth]{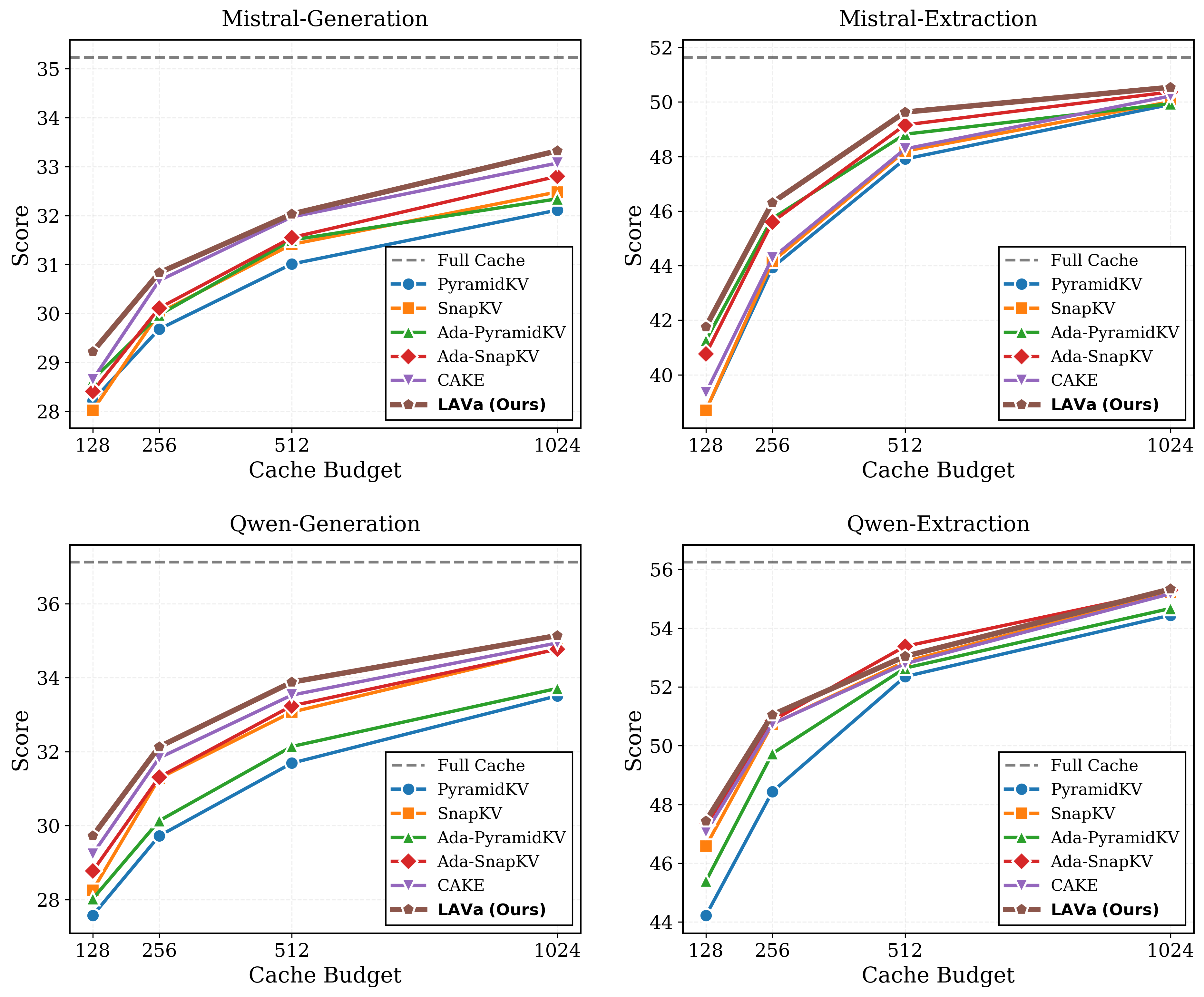}
    \caption{Results of generation and extraction tasks.}
    \label{figure:generation_extraction}
\end{figure}

To further investigate the last observation, we categorize the 20 LongBench datasets into two types: extraction tasks, which require extracting answers from the context (e.g., QA tasks evaluated with F1 or Accuracy), and generation tasks (e.g., summarization and code completion). For each category, we then compute the average scores obtained with Qwen and Mistral under varying cache budgets and eviction policies. Figure \ref{figure:generation_extraction} highlights several key findings: 1) \textit{Extraction tasks are generally less affected by compression}, as LLM performance with a compressed cache remains closer to that with a full cache; 2) \textit{The performance gap among different eviction policies is greater on generation tasks.}; 3) \textit{CAKE and LAVa outperform Ada-SnapKV and methods with fixed-layer budgets on generation tasks}, though CAKE performs significantly worse than Ada-SnapKV on extraction tasks with Mistral-7B. This suggests the importance of (dynamic) layer budget allocation for generation tasks. LAVa, however, consistently achieves top performance across both task types and language models. 

\subsection{Evaluation of Latency and Memory Peak}
\label{latency}
We evaluate LAVa's efficiency during LLM inference by analyzing peak memory usage and decoding latency on Mistral-7B-Instruct-v0.2, implemented with FlashAttention-2 \cite{dao2023flashattention}. Our comparison includes Full Cache, SnapKV, Ada-SnapKV and CAKE, all using allocation budget 1024$\mathrm{HL}$. We set input at varying lengths while keeping the output length fixed at 128.

\begin{figure}[t]
    \centering
    \hspace{-0.3cm}\includegraphics[width=1.02\linewidth]{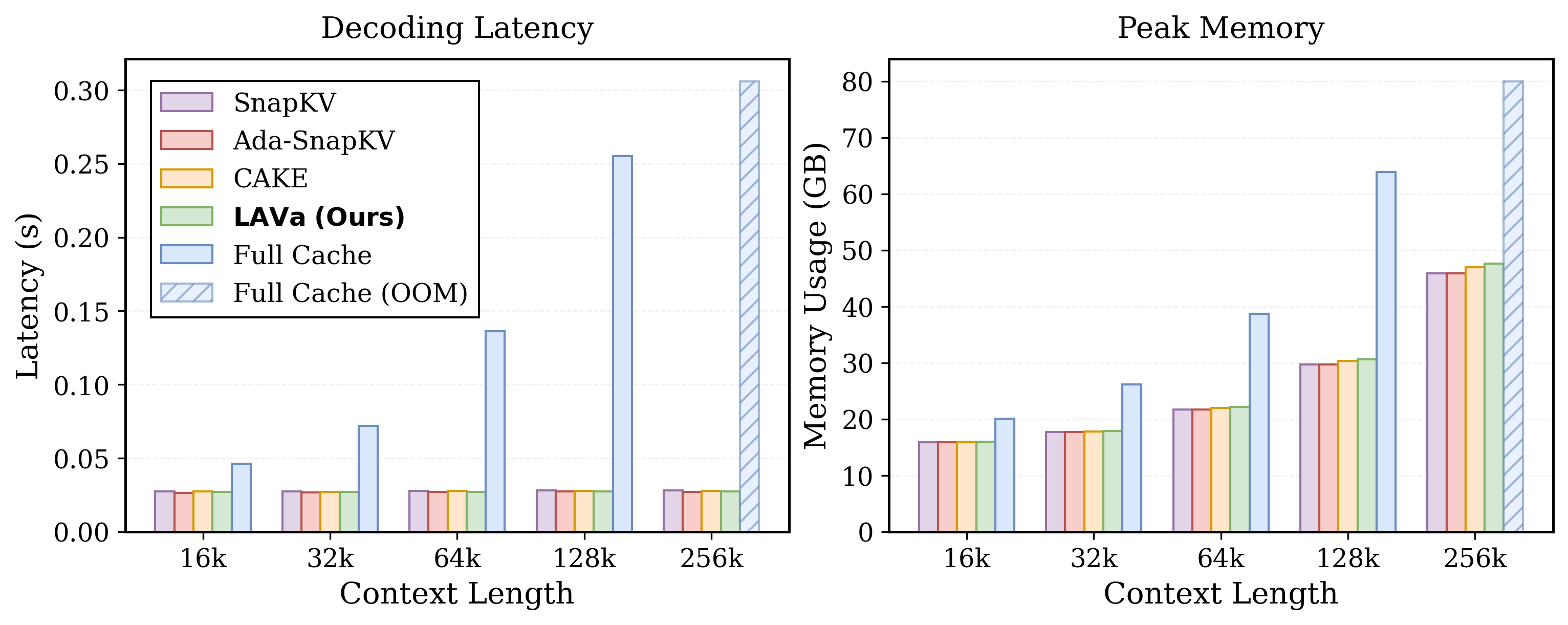}
    \caption{Peak memory usage and decoding latency in A800 80GB based on Mistral-7B-Instruct-v0.2.}
    \label{peak memory}
\end{figure}


\paragraph{Decoding Latency.}
By analyzing the decoding latency in Figure~\ref{peak memory}, we observe that our scoring function and dynamic budget allocation introduce negligible decoding cost, achieving over a 9× speedup compared to Full Cache at a 128K context length. Notably, our method is  easier to deploy than PyramidKV, Ada-PyramidKV, and CAKE, as these baselines require parameter tuning. 

\paragraph{Peak Memory Usage.}
The peak memory usage of all methods generally increases with context length due to prefilling. Our method effectively maintains peak memory at a reasonable level, particularly compared to Full Cache, which encounters OOM issues at higher context lengths. CAKE and LAVa, both employing dynamic layer budgets, generally have slightly higher peak memory usage. Compared to CAKE, LAVa requires additional storage for the norms of head-wise value vectors, but this extra memory overhead remains minimal.

{\color{black}\paragraph{Theoretical Analysis.} We provide the theoretical analysis of time complexity and memory usage in Appendix~\ref{details of experiments}. The time complexity and peak memory usage of SnapKV is $O(HN(Nd_h+wd_h+logB_{l,h}))$ and $O(HNd_h+LHB_{l,h}d_h)$, while that of LAVa is $O(HN(Nd_h+wd_h+d_h+logB_l)$ and $O(HNd_h+LHB_{l,h}d_h+LHB_{l,h}d_h)$. Setting context length $N$ as 10,000, head budget $B_{l,h}$ as 1024, the extra computation of LAVa compared to SnapKV is 0.01\% and the extra memory usage is 0.6\%, which is consistent with Figure~\ref{peak memory}.}


\subsection{Further Analysis}

\begin{figure}
    \centering
    \hspace{-0.2cm}\includegraphics[width=0.98\linewidth]{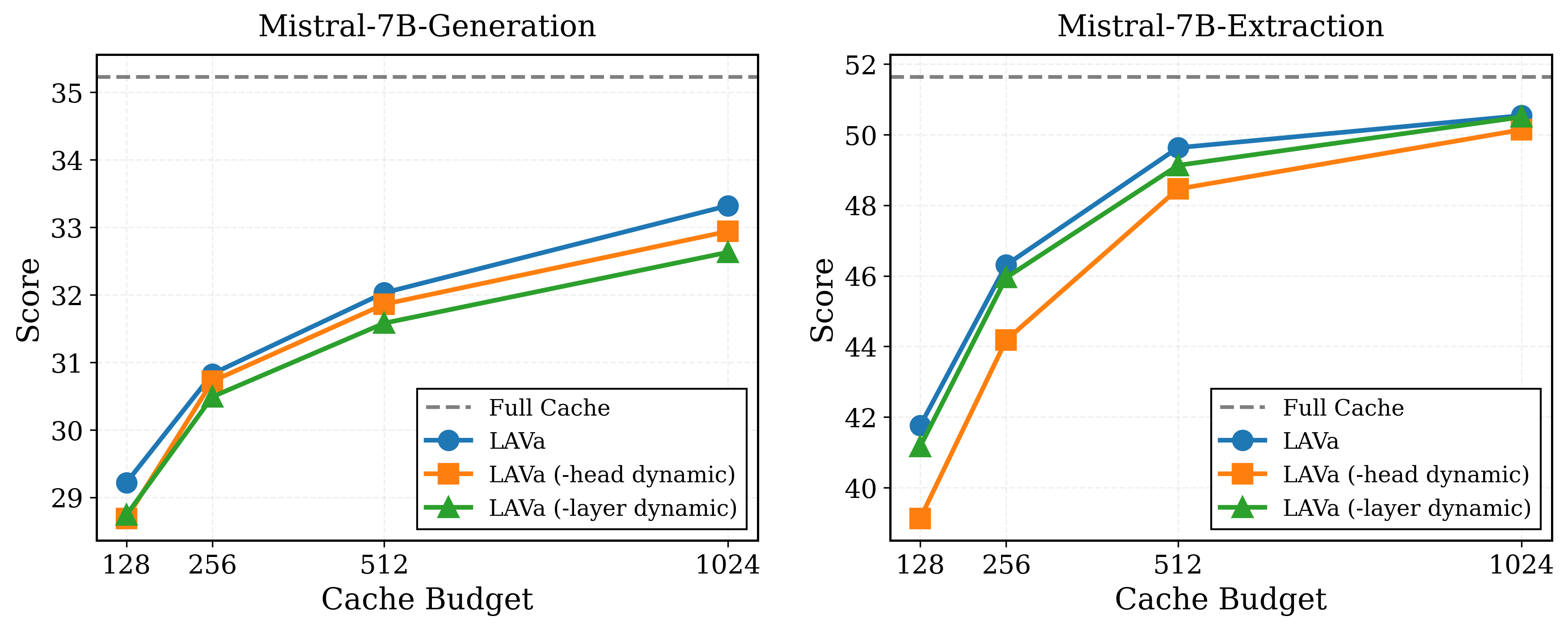}
    \caption{Ablation study on LongBench.}
    \label{fig:ab longbench}
\end{figure}

\paragraph{Dynamic Budget Allocation} To examine the impact of dynamic budget allocation, we introduce two modifications: \textbf{LAVa (-layer dynamic)}, which enforces a uniform layer budget of \(\mathbb{B}/L\), and \textbf{LAVa (-head dynamic)}, which fixes the head budget at $\mathcal{B}_l/H$ after dynamically determining the layer budget $\mathcal{B}_l$, performing head-wise cache eviction without cross-head comparisons. Results in Figure~\ref{fig:ab longbench} demonstrate that dynamic budget allocation at both the head and layer levels is essential for performance. Furthermore, it reinforces the finding that dynamic layer budgets are essential for generation tasks, whereas dynamic head budgets play a crucial role in text extraction tasks.
Detailed results are provided in Appendix~\ref{detailed ab}, where we also analyze the influence of different layer allocation approaches.

\begin{figure}
    \centering
    \hspace{-0.2cm}\includegraphics[width=0.98\linewidth]{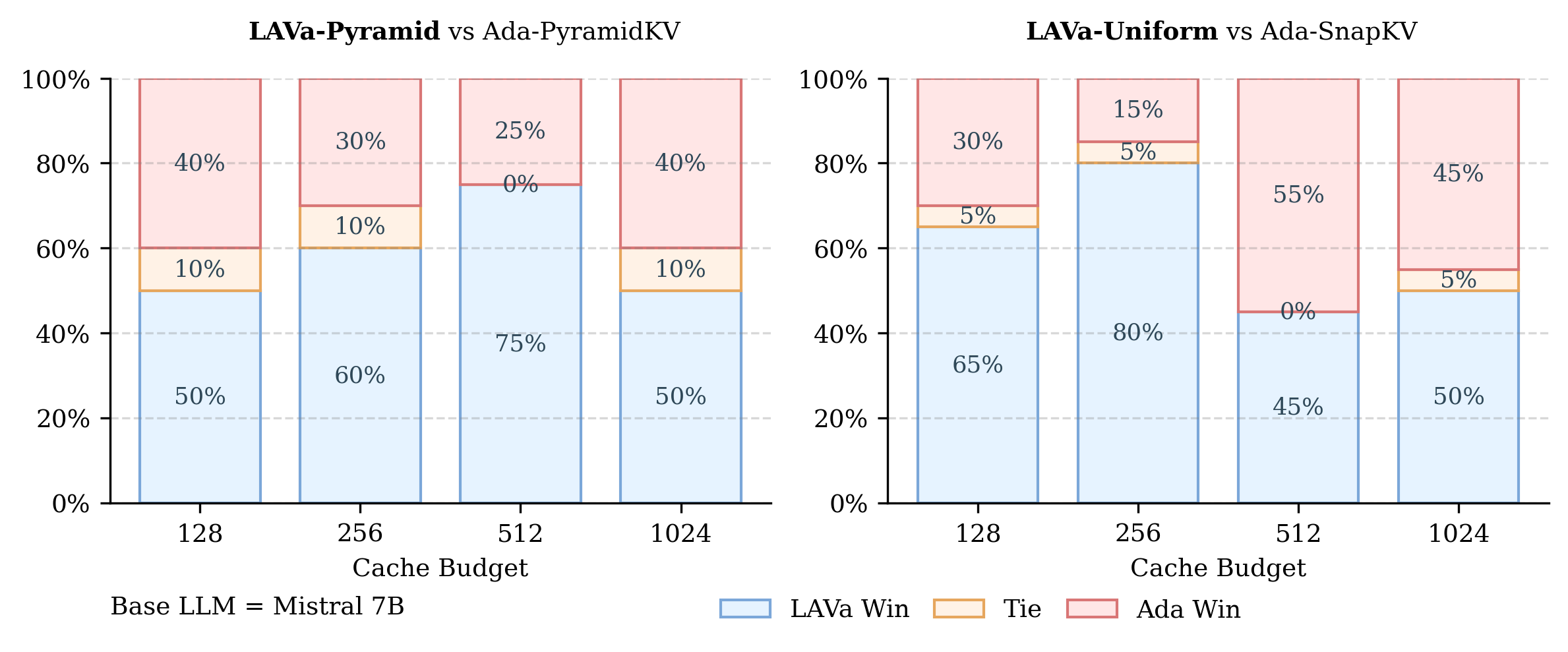}
    \caption{LaVa score vs AdaKV score on LongBench.}
    \label{lava score compare fig}
\end{figure}

\paragraph{Analysis of LAVa Score.}
To validate the effectiveness of LAVa score, we replace our dynamic layer budgets with fixed ones with PyramidKV or Uniform allocation. For different total budgets, we then compare LAVa-Pyramid with Ada-PyramidKV and LAVa-Uniform with AdaKV on LongBench. For each comparison, we count the number of tasks in LongBench where one method outperforms the other. Figure~\ref{lava score compare fig} presents the final winning rates. The results show that our scoring function yields a significantly higher number of wins in most cases, validating its effectiveness.

\section{Related Work}

Recently, various KV Cache compression methods have been proposed, leveraging different policies such as recency \cite{xiao2024efficient}, accumulated attention scores \cite{zhang2023h2o}, last-token attention scores \cite{oren2024transformers}, and recent attention scores \cite{li2024snapkv,dai2024cormcacheoptimizationrecent}. While most approaches assume a uniform budget, recent efforts have been made for dynamic budget allocation across layers \cite{qin2025cake} and heads \cite{feng2024ada}. Some methods aim at layer-dependent budgets but fix the patterns across all samples \cite{cai2024pyramidkv, yang-etal-2024-pyramidinfer}. In general, KV Cache eviction and budget allocation are typically treated as separate problems, requiring a combination of independent strategies. In contrast, we develop a principled framework based on information loss in the residual stream and propose a unified method for both cache compression and dynamic budget allocation. 

Closely related to LAVa is \cite{feng2025identifycriticalkvcache, feng2024ada}, which aims at minimizing the layer output perturbation. However, this study only applies the derived metric locally for head budget allocation. In contrast, we propose a metric for layer-wise cache eviction with dynamic layer budgets.

\section{Conclusion}

This paper provided a comprehensive of current KV Cache compression into a unified framework, grounded in the principle of minimizing information loss in Transformer residual streams. By analyzing the \textit{Layer Attention Output Loss}, we proposed LAVa, a novel layer-wise compression method that enables fully dynamic head and layer budget allocation.  Our experiments demonstrate that \textit{dynamic layer budgets are crucial for generation tasks}, whereas \textit{dynamic head budgets are important for extraction tasks}. As a fully dynamic compression method, LAVa consistently maintains top performance across task types and LLM architectures, while achieving the same speedup of $9\times$ with 128K context length compared to full cache.

Future directions include exploring new compression algorithms based on our framework, as well as extending our framework for model compression. By advancing efficient methods for LLMs, our work contributes to making LLM more accessible and scalable for diverse applications.

\section*{Limitations}
There are several limitations to our work. While we propose a unified framework with multiple optimization opportunities, our theoretical analysis and experiments focus on only one direction. Although LAVa’s simplicity is a key advantage, other approaches should be explored to further close the performance gap with a full-cache setup, particularly for generation tasks. Additionally, further research is needed to better understand why dynamic layer budget is crucial for generation tasks. Lastly, apart from FlashAttention-2~\cite{dao2023flashattention}, our method has not yet been integrated into other widely used inference frameworks, such as vLLM~\cite{10.1145/3600006.3613165}. We believe that such integration is essential for broader adoption and real-world deployment of our algorithm.

\bibliography{custom}

\begin{thebibliography}{35}
\providecommand{\natexlab}[1]{#1}

\bibitem[{Ainslie et~al.(2023)Ainslie, Lee-Thorp, de~Jong, Zemlyanskiy, Lebron, and Sanghai}]{ainslie2023gqa}
Joshua Ainslie, James Lee-Thorp, Michiel de~Jong, Yury Zemlyanskiy, Federico Lebron, and Sumit Sanghai. 2023.
\newblock \href {https://openreview.net/forum?id=hmOwOZWzYE} {{GQA}: Training generalized multi-query transformer models from multi-head checkpoints}.
\newblock In \emph{The 2023 Conference on Empirical Methods in Natural Language Processing}.

\bibitem[{Anthropic and et~al.()}]{TheC3}
Anthropic and et~al.
\newblock \href {https://api.semanticscholar.org/CorpusID:268232499} {The claude 3 model family: Opus, sonnet, haiku}.

\bibitem[{Bai et~al.(2024)Bai, Lv, Zhang, Lyu, Tang, Huang, Du, Liu, Zeng, Hou, Dong, Tang, and Li}]{bai-etal-2024-longbench}
Yushi Bai, Xin Lv, Jiajie Zhang, Hongchang Lyu, Jiankai Tang, Zhidian Huang, Zhengxiao Du, Xiao Liu, Aohan Zeng, Lei Hou, Yuxiao Dong, Jie Tang, and Juanzi Li. 2024.
\newblock \href {https://doi.org/10.18653/v1/2024.acl-long.172} {{L}ong{B}ench: A bilingual, multitask benchmark for long context understanding}.
\newblock In \emph{Proceedings of the 62nd Annual Meeting of the Association for Computational Linguistics (Volume 1: Long Papers)}, pages 3119--3137, Bangkok, Thailand. Association for Computational Linguistics.

\bibitem[{Cai et~al.(2024)Cai, Zhang, Gao, Liu, Liu, Lu, Xiong, Dong, Chang, Hu et~al.}]{cai2024pyramidkv}
Zefan Cai, Yichi Zhang, Bofei Gao, Yuliang Liu, Tianyu Liu, Keming Lu, Wayne Xiong, Yue Dong, Baobao Chang, Junjie Hu, et~al. 2024.
\newblock Pyramidkv: Dynamic kv cache compression based on pyramidal information funneling.
\newblock \emph{arXiv preprint arXiv:2406.02069}.

\bibitem[{Chen et~al.(2025)Chen, Li, Shieh, and Bing}]{chen2025longpo}
Guanzheng Chen, Xin Li, Michael Shieh, and Lidong Bing. 2025.
\newblock \href {https://openreview.net/forum?id=qTrEq31Shm} {Long{PO}: Long context self-evolution of large language models through short-to-long preference optimization}.
\newblock In \emph{The Thirteenth International Conference on Learning Representations}.

\bibitem[{Chiang et~al.(2023)Chiang, Li, Lin, Sheng, Wu, Zhang, Zheng, Zhuang, Zhuang, Gonzalez, Stoica, and Xing}]{vicuna2023}
Wei-Lin Chiang, Zhuohan Li, Zi~Lin, Ying Sheng, Zhanghao Wu, Hao Zhang, Lianmin Zheng, Siyuan Zhuang, Yonghao Zhuang, Joseph~E. Gonzalez, Ion Stoica, and Eric~P. Xing. 2023.
\newblock \href {https://lmsys.org/blog/2023-03-30-vicuna/} {Vicuna: An open-source chatbot impressing gpt-4 with 90\%* chatgpt quality}.

\bibitem[{Dai et~al.(2024)Dai, Huang, Jiang, Chen, Cai, Bi, and Shi}]{dai2024cormcacheoptimizationrecent}
Jincheng Dai, Zhuowei Huang, Haiyun Jiang, Chen Chen, Deng Cai, Wei Bi, and Shuming Shi. 2024.
\newblock \href {https://arxiv.org/abs/2404.15949} {Corm: Cache optimization with recent message for large language model inference}.
\newblock \emph{Preprint}, arXiv:2404.15949.

\bibitem[{Dao(2023)}]{dao2023flashattention}
Tri Dao. 2023.
\newblock Flashattention-2: Faster attention with better parallelism and work partitioning.
\newblock \emph{arXiv preprint arXiv:2307.08691}.

\bibitem[{Feng et~al.(2024)Feng, Lv, Cao, Xie, and Zhou}]{feng2024ada}
Yuan Feng, Junlin Lv, Yukun Cao, Xike Xie, and S~Kevin Zhou. 2024.
\newblock Ada-kv: Optimizing kv cache eviction by adaptive budget allocation for efficient llm inference.
\newblock \emph{arXiv preprint arXiv:2407.11550}.

\bibitem[{Feng et~al.(2025)Feng, Lv, Cao, Xie, and Zhou}]{feng2025identifycriticalkvcache}
Yuan Feng, Junlin Lv, Yukun Cao, Xike Xie, and S~Kevin Zhou. 2025.
\newblock \href {https://arxiv.org/abs/2502.03805} {Identify critical kv cache in llm inference from an output perturbation perspective}.
\newblock \emph{Preprint}, arXiv:2502.03805.

\bibitem[{Ferrando and Voita(2024)}]{ferrando-voita-2024-information}
Javier Ferrando and Elena Voita. 2024.
\newblock \href {https://doi.org/10.18653/v1/2024.emnlp-main.965} {Information flow routes: Automatically interpreting language models at scale}.
\newblock In \emph{Proceedings of the 2024 Conference on Empirical Methods in Natural Language Processing}, pages 17432--17445, Miami, Florida, USA. Association for Computational Linguistics.

\bibitem[{Fu et~al.(2024)Fu, Panda, Niu, Yue, Hajishirzi, Kim, and Peng}]{needle2}
Yao Fu, Rameswar Panda, Xinyao Niu, Xiang Yue, Hannaneh Hajishirzi, Yoon Kim, and Hao Peng. 2024.
\newblock Data engineering for scaling language models to 128k context.
\newblock In \emph{Proceedings of the 41st International Conference on Machine Learning}, ICML'24. JMLR.org.

\bibitem[{Guo et~al.(2023)Guo, Xu, Duan, Yin, and McAuley}]{Guo2023LongCoderAL}
Daya Guo, Canwen Xu, Nan Duan, Jian Yin, and Julian McAuley. 2023.
\newblock \href {https://api.semanticscholar.org/CorpusID:259262301} {Longcoder: A long-range pre-trained language model for code completion}.
\newblock In \emph{International Conference on Machine Learning}.

\bibitem[{Guo et~al.(2024)Guo, Kamigaito, and Watanabe}]{guo-etal-2024-attention}
Zhiyu Guo, Hidetaka Kamigaito, and Taro Watanabe. 2024.
\newblock \href {https://doi.org/10.18653/v1/2024.emnlp-main.1178} {Attention score is not all you need for token importance indicator in {KV} cache reduction: Value also matters}.
\newblock In \emph{Proceedings of the 2024 Conference on Empirical Methods in Natural Language Processing}, pages 21158--21166, Miami, Florida, USA. Association for Computational Linguistics.

\bibitem[{Horn and Johnson(2012)}]{horn2012matrix}
Roger~A Horn and Charles~R Johnson. 2012.
\newblock \emph{Matrix analysis}.
\newblock Cambridge university press.

\bibitem[{Hsieh et~al.(2024)Hsieh, Sun, Kriman, Acharya, Rekesh, Jia, and Ginsburg}]{hsieh2024ruler}
Cheng-Ping Hsieh, Simeng Sun, Samuel Kriman, Shantanu Acharya, Dima Rekesh, Fei Jia, and Boris Ginsburg. 2024.
\newblock \href {https://openreview.net/forum?id=kIoBbc76Sy} {{RULER}: What{\textquoteright}s the real context size of your long-context language models?}
\newblock In \emph{First Conference on Language Modeling}.

\bibitem[{Hu et~al.(2021)Hu, Shen, Wallis, Allen-Zhu, Li, Wang, Wang, and Chen}]{hu2021lora}
Edward~J Hu, Yelong Shen, Phillip Wallis, Zeyuan Allen-Zhu, Yuanzhi Li, Shean Wang, Lu~Wang, and Weizhu Chen. 2021.
\newblock Lora: Low-rank adaptation of large language models.
\newblock \emph{arXiv preprint arXiv:2106.09685}.

\bibitem[{Jiang et~al.(2023)Jiang, Sablayrolles, Mensch, Bamford, Chaplot, de~las Casas, Bressand, Lengyel, Lample, Saulnier, Lavaud, Lachaux, Stock, Scao, Lavril, Wang, Lacroix, and Sayed}]{jiang2023mistral7b}
Albert~Q. Jiang, Alexandre Sablayrolles, Arthur Mensch, Chris Bamford, Devendra~Singh Chaplot, Diego de~las Casas, Florian Bressand, Gianna Lengyel, Guillaume Lample, Lucile Saulnier, Lélio~Renard Lavaud, Marie-Anne Lachaux, Pierre Stock, Teven~Le Scao, Thibaut Lavril, Thomas Wang, Timothée Lacroix, and William~El Sayed. 2023.
\newblock \href {https://arxiv.org/abs/2310.06825} {Mistral 7b}.
\newblock \emph{Preprint}, arXiv:2310.06825.

\bibitem[{Kamalloo et~al.(2023)Kamalloo, Dziri, Clarke, and Rafiei}]{kamalloo-etal-2023-evaluating}
Ehsan Kamalloo, Nouha Dziri, Charles Clarke, and Davood Rafiei. 2023.
\newblock \href {https://doi.org/10.18653/v1/2023.acl-long.307} {Evaluating open-domain question answering in the era of large language models}.
\newblock In \emph{Proceedings of the 61st Annual Meeting of the Association for Computational Linguistics (Volume 1: Long Papers)}, pages 5591--5606, Toronto, Canada. Association for Computational Linguistics.

\bibitem[{Kwon et~al.(2023)Kwon, Li, Zhuang, Sheng, Zheng, Yu, Gonzalez, Zhang, and Stoica}]{10.1145/3600006.3613165}
Woosuk Kwon, Zhuohan Li, Siyuan Zhuang, Ying Sheng, Lianmin Zheng, Cody~Hao Yu, Joseph Gonzalez, Hao Zhang, and Ion Stoica. 2023.
\newblock \href {https://doi.org/10.1145/3600006.3613165} {Efficient memory management for large language model serving with pagedattention}.
\newblock In \emph{Proceedings of the 29th Symposium on Operating Systems Principles}, SOSP '23, page 611–626, New York, NY, USA. Association for Computing Machinery.

\bibitem[{Li et~al.(2024)Li, Huang, Yang, Venkitesh, Locatelli, Ye, Cai, Lewis, and Chen}]{li2024snapkv}
Yuhong Li, Yingbing Huang, Bowen Yang, Bharat Venkitesh, Acyr Locatelli, Hanchen Ye, Tianle Cai, Patrick Lewis, and Deming Chen. 2024.
\newblock \href {https://openreview.net/forum?id=poE54GOq2l} {Snap{KV}: {LLM} knows what you are looking for before generation}.
\newblock In \emph{The Thirty-eighth Annual Conference on Neural Information Processing Systems}.

\bibitem[{Liu et~al.(2024)Liu, Lin, Hewitt, Paranjape, Bevilacqua, Petroni, and Liang}]{liu-etal-2024-lost}
Nelson~F. Liu, Kevin Lin, John Hewitt, Ashwin Paranjape, Michele Bevilacqua, Fabio Petroni, and Percy Liang. 2024.
\newblock \href {https://doi.org/10.1162/tacl_a_00638} {Lost in the middle: How language models use long contexts}.
\newblock \emph{Transactions of the Association for Computational Linguistics}, 12:157--173.

\bibitem[{OpenAI and et~al.(2024)}]{openai2024gpt4technicalreport}
OpenAI and et~al. 2024.
\newblock \href {https://arxiv.org/abs/2303.08774} {Gpt-4 technical report}.
\newblock \emph{Preprint}, arXiv:2303.08774.

\bibitem[{Oren et~al.(2024)Oren, Hassid, Yarden, Adi, and Schwartz}]{oren2024transformers}
Matanel Oren, Michael Hassid, Nir Yarden, Yossi Adi, and Roy Schwartz. 2024.
\newblock Transformers are multi-state rnns.
\newblock \emph{arXiv preprint arXiv:2401.06104}.

\bibitem[{Qin et~al.(2024{\natexlab{a}})Qin, Rosset, Chau, Rao, and Van~Durme}]{qin2024dodo}
Guanghui Qin, Corby Rosset, Ethan Chau, Nikhil Rao, and Benjamin Van~Durme. 2024{\natexlab{a}}.
\newblock Dodo: Dynamic contextual compression for decoder-only lms.
\newblock In \emph{Proceedings of the 62nd Annual Meeting of the Association for Computational Linguistics (Volume 1: Long Papers)}, pages 9961--9975.

\bibitem[{Qin et~al.(2024{\natexlab{b}})Qin, Rosset, Chau, Rao, and Durme}]{qin2024nugget}
Guanghui Qin, Corby Rosset, Ethan~C. Chau, Nikhil Rao, and Benjamin~Van Durme. 2024{\natexlab{b}}.
\newblock \href {https://openreview.net/forum?id=jVsXDLIt45} {Nugget 2d: Dynamic contextual compression for scaling decoder-only language models}.

\bibitem[{Qin et~al.(2025)Qin, Cao, Lin, Hu, Fan, Cheng, Lin, and Li}]{qin2025cake}
Ziran Qin, Yuchen Cao, Mingbao Lin, Wen Hu, Shixuan Fan, Ke~Cheng, Weiyao Lin, and Jianguo Li. 2025.
\newblock \href {https://openreview.net/forum?id=EQgEMAD4kv} {{CAKE}: Cascading and adaptive {KV} cache eviction with layer preferences}.
\newblock In \emph{The Thirteenth International Conference on Learning Representations}.

\bibitem[{Qwen and et~al.(2025)}]{qwen2025qwen25technicalreport}
Qwen and et~al. 2025.
\newblock \href {https://arxiv.org/abs/2412.15115} {Qwen2.5 technical report}.
\newblock \emph{Preprint}, arXiv:2412.15115.

\bibitem[{Vaswani(2017)}]{vaswani2017attention}
A~Vaswani. 2017.
\newblock Attention is all you need.
\newblock \emph{Advances in Neural Information Processing Systems}.

\bibitem[{Wu et~al.(2025)Wu, Wang, Xiao, Peng, and Fu}]{wu2024retrievalheadmechanisticallyexplains}
Wenhao Wu, Yizhong Wang, Guangxuan Xiao, Hao Peng, and Yao Fu. 2025.
\newblock \href {https://openreview.net/forum?id=EytBpUGB1Z} {Retrieval head mechanistically explains long-context factuality}.
\newblock In \emph{The Thirteenth International Conference on Learning Representations}.

\bibitem[{Xiao et~al.(2025)Xiao, Tang, Zuo, junxian guo, Yang, Tang, Fu, and Han}]{xiao2025duoattention}
Guangxuan Xiao, Jiaming Tang, Jingwei Zuo, junxian guo, Shang Yang, Haotian Tang, Yao Fu, and Song Han. 2025.
\newblock \href {https://openreview.net/forum?id=cFu7ze7xUm} {Duoattention: Efficient long-context {LLM} inference with retrieval and streaming heads}.
\newblock In \emph{The Thirteenth International Conference on Learning Representations}.

\bibitem[{Xiao et~al.(2024)Xiao, Tian, Chen, Han, and Lewis}]{xiao2024efficient}
Guangxuan Xiao, Yuandong Tian, Beidi Chen, Song Han, and Mike Lewis. 2024.
\newblock \href {https://openreview.net/forum?id=NG7sS51zVF} {Efficient streaming language models with attention sinks}.
\newblock In \emph{The Twelfth International Conference on Learning Representations}.

\bibitem[{Yang et~al.(2024)Yang, Han, Gao, Hu, Zhang, and Zhao}]{yang-etal-2024-pyramidinfer}
Dongjie Yang, Xiaodong Han, Yan Gao, Yao Hu, Shilin Zhang, and Hai Zhao. 2024.
\newblock \href {https://doi.org/10.18653/v1/2024.findings-acl.195} {{P}yramid{I}nfer: Pyramid {KV} cache compression for high-throughput {LLM} inference}.
\newblock In \emph{Findings of the Association for Computational Linguistics: ACL 2024}, pages 3258--3270, Bangkok, Thailand. Association for Computational Linguistics.

\bibitem[{Zhang et~al.(2024)Zhang, Chen, Hu, Xu, Chen, Hao, Han, Thai, Wang, Liu, and Sun}]{zhang-etal-2024-bench}
Xinrong Zhang, Yingfa Chen, Shengding Hu, Zihang Xu, Junhao Chen, Moo Hao, Xu~Han, Zhen Thai, Shuo Wang, Zhiyuan Liu, and Maosong Sun. 2024.
\newblock \href {https://doi.org/10.18653/v1/2024.acl-long.814} {$\infty${B}ench: Extending long context evaluation beyond 100{K} tokens}.
\newblock In \emph{Proceedings of the 62nd Annual Meeting of the Association for Computational Linguistics (Volume 1: Long Papers)}, pages 15262--15277, Bangkok, Thailand. Association for Computational Linguistics.

\bibitem[{Zhang et~al.(2023)Zhang, Sheng, Zhou, Chen, Zheng, Cai, Song, Tian, R{\'e}, Barrett et~al.}]{zhang2023h2o}
Zhenyu Zhang, Ying Sheng, Tianyi Zhou, Tianlong Chen, Lianmin Zheng, Ruisi Cai, Zhao Song, Yuandong Tian, Christopher R{\'e}, Clark Barrett, et~al. 2023.
\newblock H2o: Heavy-hitter oracle for efficient generative inference of large language models.
\newblock \emph{Advances in Neural Information Processing Systems}, pages 34661--34710.

\end{thebibliography}

\appendix

\section{Extension of The Information Flow of LLM Decoding Process with KV Cache}
KV Cache is initialized at \textit{prefilling} stage, which basically computes the Key and Value for tokens in the initial prompts in the standard way \cite{vaswani2017attention}. In the following, we assume that there exists a KV Cache of $(N-1)$ previous tokens and demonstrate how decoding is performed at step-$N$.


\paragraph{Notation Table}
\label{ref:notation}The LLM has $L$ layers, each has $H$ heads. The model and head dimensions are $d$ and $d_h=d/H$; $K_l, V_l$ are the KV Cache for the $l$-th layer up to the current time step (the $N$-th token), which are of $[H, (N-1), d_h]$ sizes. The notations for the theoretical analysis are listed in Table~\ref{notation table}. 

\begin{table*}[t]
    \centering
    \begin{adjustbox}{max width=\textwidth}
    \begin{tabular}{c|c|c|c} \hline
    \toprule
        \textbf{Notation} & \textbf{Explanation} & \textbf{Notation} & \textbf{Explanation} \\\midrule
        $N$ & Current token length & $A_{l,h}^N[i]$ &  Attention weight of position $i$ at layer $l$, head $h$ and step $N$ \\\hline
        $N_e$ & Expected token length  & $y_l^N$ & Attention output of layer $l$ and step $N$ \\\hline
        $L$ & Total number of layers & $\hat{y}_l^N$ & Modified attention output of layer $l$ and step $N$ after eviction \\\hline
        $H$ & Total number of heads per layer & $p$ & Logits after last layer for next token\\\hline
        $l$ & Layer index, $l\in [L]$ & $\hat{p}$ & Modified logits after last layer for next token after eviction\\\hline
        $h$ & Head index, $h\in[H]$ & $\mathcal{P}$ & Information loss function of Transformer residual streams \\\hline
        $d$ & The model embedding dimension & $w$ & Sliding window size\\\hline
        $d_h$ & The head embedding dimension $d_h=d/H$ & $\mathcal{B}_{l,h}$ & Budget for head $h$ of layer $l$\\\hline
        $x_l^N$ & The input hidden states of step $N$ and layer $l$ & $\mathcal{B}_l$ &  Budget for layer $l$ \\\hline
        $Q_l^N$ & The query vector of step $N$ and layer $l$ & $\mathbb{B}$ & Fixed total budget for KV Cache, $\mathbb{B}=\sum_{l\in[L]}\mathcal{B}_l$\\\hline
        $K_l^N$ & The key vector of step $N$ and layer $l$ &          $s_{l,h}[i]$ & Score of position $i$ at layer $l$ and head $h$
\\\hline
        $V_l^N$ & The value vector of step $N$ and layer $l$ & $ e_l$ & The uncertainty of layer $l$ for dynamic layer budget allocation\\\hline
        $K_{l,h}$ & Key cache of layer $l$ and head $h$ & $\mathcal{I}_{l,h}$ & Attention mask for the head $h$ of layer $l$, $\mathcal{I}_{l,h}\in [1,0]^{N}$\\\hline
        $V_{l,h}$ & Value cache of layer $l$ and head $h$ & $\mathcal{I}$ & Attention mask $\mathcal{I}\in[1,0]^{L\times H \times N}$ \\
    \bottomrule
    \end{tabular}
    \end{adjustbox}
    \caption{Notation table.}
    \label{notation table}
\end{table*}

\paragraph{Decoding Process} According to \cite{ferrando-voita-2024-information}, the decoding process of large language models (LLMs) can be viewed as a series of operations on the current \textit{residual stream}, as illustrated in Figure \ref{fig:transformer-residual-streams}. In each layer, information is read from the residual stream, updated, and then written back. Specifically, supposing that $x^N_l$ is the current input for layer $l$, we first calculate the corresponding $Q^N_l, K^N_l, V^N_l$ as follows:
\begin{equation}
    Q^N_l = x^N_lW_l^Q; K^N_l = x^N_lW_l^K; V^N_l = x^N_lW_l^V \nonumber
\end{equation}
where $Q^N_l, K^N_l, V^N_l$ are of size ($H\times 1 \times d_h$), containing $H$ head-wise caches. 
The layer-wise KV Cache is then updated as follows:
\begin{equation}
    K_l = Cat[K_l,K^N_l], V_l = Cat[V_l,V^N_l] \nonumber
\end{equation}
where $K_l, V_l$ are tensors of size ($H\times N\times d_h$), and $Cat$ indicates the concatenation operation.
We then calculate the attention scores of step-$N$ for layer-$l$:
\begin{equation}
    A^N_l = Cat_{h\in[H]}\left(A^N_{l,h}\right)   \notag
\end{equation}
where $A^N_{l,h}=Softmax(\frac{Q^N_{l,h}{K_{l,h}}}{\sqrt{d_h}})$. Here, {textbf{\textit{$A^N_{l,h}[i]$ indicates how much the token at step-$N$ (the $N$-th token) attends to the $i$-th token ($i<=N$)}}. Layer-$l$ attention output is calculated as follows:
\begin{align}
    &y^N_l = Cat_{h\in[H]}(A_{l,h}^NV_{l,h} )W_l^O \in \mathbb{R}^{1\times d} \notag
\end{align}
where $W_l^O\in \mathbb{R}^{d\times d}$. The layer output $x^N_{l+1}$, which is also the input for the layer-$(l+1)$, is calculated as $x^N_{l+1}=y^N_l+FFN(y^N_l)$.

In the last layer, we exploits an un-embedding layer ($W^M\in\mathbb{R}^{d\times |\mathcal{V}|}$) to get the probability vector $p$ for next token sampling:

\begin{align}
    p^N = \left(y^N_L+FFN(y^N_L\right) W^M 
\end{align}

\paragraph{Head-wise vs Layer-wise Cache}
\label{head and layer cache}
Current query matrix and KV Cache on head $h$ of layer $l$ are :
\begin{equation}
    Q_{l,h}^N = Q_l^N[:,d_h*h:d_h*(h+1)] \in \mathbb{R}^{1\times d_h}
\end{equation}
\begin{align}
        K_{l,h} &= K_l[:,d_h*h:d_h*(h+1)], \\    V_{l,h} &= V_l[:,d_h*h:d_h*(h+1)] \in \mathbb{R}^{N\times d_h}
\end{align}
Henc, the layer-wise KV Cache can be treated as concatenation of head-wise elements where we just change the order of dimensions:
\begin{align}
    K_l &= Cat_{h\in[H]}[K_{l,h}] \in \mathbb{R}^{H\times N \times d_h},
    \label{layer-wise k}\\
    V_l &= Cat_{h\in[H]}[V_{l,h}] \in \mathbb{R}^{H\times N \times d_h}
\label{layer-wise v}
\end{align}
And the same to the query matrix:
\begin{equation}
    Q_l^N = Cat_{h\in[H]}[Q^N_{l,h}]\in\mathbb{R}^{H\times 1\times d_h}
\label{layer-wise q}
\end{equation}

\section{Extension of A Principled Framework for KV Cache Eviction based on Information Loss} 

The unified problem for budget allocation and cache eviction can be defined as follows:
\begin{align}
    &\min_{\mathcal{I},\mathcal{B}} \mathcal{P}(\textbf{x}^{1\ldots N}_1,\mathcal{I},\mathcal{B}) 
        \label{loss objective appendix}\\
    \text{st.} &\sum_{i\in[N]}\mathcal{I}_{l,h}[i]=\mathcal{B}_{l,h}; \nonumber \\
    &\sum_{h\in[H]}\mathcal{B}_{l,h}=\mathcal{B}_l; 
    \sum_{l\in[L]}\mathcal{B}_{l} = \mathbb{B} \nonumber\\ 
    &\mathcal{I}_{l,h}[k] = 1 \text{ },\forall l,h; \text{ and } \forall k \in [N-w,N]\nonumber
\end{align}
The optimization problem in Eq.~\ref{loss objective appendix} is infeasible to solve for several reasons. We can instead search for a scoring function \(s\), where \(s_{l,h}[i]\) assigns an importance score to token \(i\) at layer \(l\) and head \(h\). This scoring function allows us to greedily choose the least important entries to be masked until the budget is met $\mathcal{I}=Select(s, \mathcal{B})$. Bringing everything together, we arrive at the following (surrogate) optimization problem:
\begin{align}
    &\min_{\mathcal{B}, s\in\mathcal{F}} \mathcal{P}(\textbf{x}^{1\ldots N}_1,s,\mathbb{B}) 
        \label{loss objective1 appendix}
\end{align}

\label{descriptions}

\begin{table*}[t]
\centering
\small
\begin{tabular}{@{}l|c|c|p{5.5cm}|p{2cm}|}
\toprule
\multirow{2}{*}{\textbf{Methods}} & \multicolumn{2}{|c|}{\textbf{Budgets}} & \multirow{2}{*}{\textbf{Scoring Function}} & \multirow{2}{*}{\textbf{Loss}} \\ \cmidrule{2-3}
& $\mathcal{B}_{l,h}$ & $\mathcal{B}_{l}$ & & \\\cmidrule{1-5}
H2O \cite{zhang2023h2o} & $\mathcal{B}_l/H$ & $\mathbb{B}/L$ & Accumulated attention scores &\multirow{9}{=}{Head Attention} \\
 & & &$s_{l,h}[i]=\sum_{j=i+1}^NA^j_{l,h}[i]$ & \\
SnapKV \cite{li2024snapkv} & $\mathcal{B}_l/H$ & $\mathbb{B}/L$ & Recent attention scores & \\
 & & &   $s_{l,h}[i]=\frac{1}{w}\sum_{j=N-w}^NA_{l,h}^j[i]$, $\forall i<N-w$ & \\ 
TOVA \cite{oren2024transformers} & $\mathcal{B}_l/H$ & $\mathbb{B}/L$ & Last-token attention scores  & \\
& & & $s_{l,h}[i]=A^N_{l,h}[i]$  & \\
CAKE \cite{qin2025cake} & $\mathcal{B}_l/H$ & Dynamic & Recent attention scores + attention shifts & \\
& & & $s_{l,h}[i]= \gamma \text{VAR}_{j=N-w}^N([A_{l,h}^j[i]))$ & \\
& & & $+\frac{1}{w}\sum_{j=N-w}^N A_{l,h}^{j}[i] $, $\forall i<N-w$  & \\
\cmidrule{1-5}
VATP \cite{guo-etal-2024-attention} & $\mathcal{B}_l/H$ & $\mathbb{B}/L$ & Recent attention scores + value vectors & \multirow{2}{=}{Head Attention Output} \\
& & &   $s_{l,h}[i]=\frac{\|V_{l,h}[i]\|_1}{w}\sum_{j=N-w}^NA_{l,h}^j[i]$ & \\ 
\cmidrule{1-5}
Dodo \cite{qin2024dodo} & Dynamic & $\mathbb{B}/L$ & Neural Network (LoRA) & Logits  \\
\cmidrule{1-5}
DuoAttention \cite{xiao2025duoattention} & $w$ or full & - & Head classifier (retrieval vs non-retrieval) & \multirow{4}{=}{Layer Attention Output}  \\
AdaKV \cite{feng2024ada} & Dynamic & Fixed & Recent attention scores \\
\textbf{LAVa (Ours)} & Dynamic & Dynamic & Recent attention scores + value vectors & \\
& & & $s_{l,h}[i]=\frac{max_{k}\|V_{l,h}[k]\|_1}{w}\sum_{j=N-w}^NA_{l,h}^j[i]$ &\\
\bottomrule
\end{tabular}
\caption{Comparison between different methods; Dodo and DuoAttention require training; The layer cache budget $\mathcal{B}_l$ of AdaKV is based on the method it is integrated with.}
\label{tab:summary full}
\end{table*}

Current various kv cache eviction methods can be adapted into our framework, just defining several significant functions and parameters (including $P, \mathcal{I}, \mathcal{B}$ and $s$) and introducing additional constraints, which will result in suboptimal performance. In addition, they adopt many heuristic techniques based on observations to simplify the problem. The full summarization of how existeing methods can be formalized within our framework is presented in Table~\ref{tab:summary full}.


\paragraph{H2O.} \cite{zhang2023h2o}
Allocation budgets $\mathcal{B}$ are all fixed before generation. The budgets of all layers are the same and the budgets of all heads are also the same.
\begin{equation}
    \mathcal{B}_{l,h} = \frac{\mathcal{B}}{HL} 
\end{equation}
H2O uses \textbf{head attention loss} and adopt accumulated attention scores as score function.
\begin{equation}
    s_{l,h}[i]=\sum_{j=i+1}^{N} A_{l,h}^j[i] , \mathcal{I}_{l,h}=Select(s_{l,h},\mathcal{B}_{l,h})
\end{equation}
\noindent H2O claimed that the accumulated attention score can preserve the future attention pattern better. This technique is heuristic and based on observations of experiments in several methods like H2O and SnapKV \cite{li2024snapkv}, but it is valid and actually can improve the performance, mitigating the impact of absolutism of only current attention scores \cite{oren2024transformers}.

\paragraph{TOVA.} \cite{oren2024transformers}
The difference between TOVA and H2O is that TOVA uses current attention scores as score function. 
\begin{equation}
    s_{l,h}[i]=A_{l,h}^N[i], \mathcal{I}_{l,h}=Select(s_{l,h},\mathcal{B}_{l,h})
\end{equation}

\paragraph{SnapKV.} \cite{li2024snapkv}
The difference between SnapKV and H2O is that SnapKV uses recent attention scores as score function, which means SnapKV only utilizes tokens within sliding window to calculate accumulated attention scores. We set sliding window size as $w$:
\begin{align}
    &s_{l,h}[i]=\sum_{j=N-w}^N A_{l,h}^j[i] \notag\\ 
    &\mathcal{I}_{l,h}=Select(s_{l,h},\mathcal{B}_{l,h})
    \label{accumulated and pooling attention scores}
\end{align}
SnapKV claims that the accumulated attention scores of the recent sliding window is enough to represent the significance of tokens. Furthermore, SnapKV adopts pooling operation to preserve the completeness of the information. In our view, better protecting the coherence of the text is the reason for the effectiveness of pooling operation.

\paragraph{PyramidKV.} \cite{cai2024pyramidkv}
The difference between PyramidKV and SnapKV is that considering the different significance of layers in the long-context setting, PyramidKV set the budgets of layers in a descending order like a pyramid. It uses a hyper-parameter $\beta$ to control the shape of pyramid.
\begin{align}
    \mathcal{B}_{L-1}=\frac{\mathcal{B}}{\beta*L} &, \mathcal{B}_{0}=\frac{2*\mathcal{B}}{L}-\mathcal{B}_{L-1} \notag \\
    \mathcal{B}_{l} = \mathcal{B}_{0} &- \frac{\mathcal{B}_{L-1}-\mathcal{B}_{0}}{L-1}*l
\end{align}
And the budgets of heads in one layer are the same: $\mathcal{B}_{l,h}=\frac{\mathcal{B}_{l}}{H}$.

Hence, compared with SnapKV, PyramidKV consider about different budgets of layers in a heuristic way.
\paragraph{CAKE.} \citep{qin2025cake}
Allocation budgets $\mathcal{B}$ are generated through the online prefilling stage. All heads of one layer have the same budget. So CAKE do not consider the level of head (such as using mean information across heads).

Considering spatial and temporal information, CAKE allocates different budgets to different layers. And not adopting the fixed pattern like PyramidKV, CAKE claims that for different samples, the allocation pattern also needs to be adapted. It defines functions of spatial and temporal information for one layer $l$, the spatial information function $\mathcal{H}$ is formed as entropy of attention scores (larger values means more even distribution) and the temporal information function $\mathcal{V}$ (larger values means more distribution shift) is formed as variance of attention scores ($A^{(n)}$ means the attention scores distribution in the n-th step of prefilling stage):
\begin{align}
    \mathcal{H}_l &= -\sum^{N}_{j=1}A_{l}^{j}\log(A_{l}^{j}), \notag     \\ \mathcal{V}_l &= \sum^{N}_{j=1}\text{VAR}([A_l^{t}[j]]^{t\in[j,N]})
\end{align}
Then CAKE uses these two functions to determine the budget of layers, where $\gamma_1$ and $\gamma_2$ are two hyper-parameters to control the influence of two functions:
\begin{equation}
    \mathcal{P}_l = \mathcal{H}_l^{\frac{1}{\gamma_1}}\mathcal{V}_l^{\frac{1}{\gamma_2}} , \mathcal{B}_l = \frac{\mathcal{P}_l}{\sum^{l\in[L]}\mathcal{P}_l}\mathcal{B}, \mathcal{B}_{l,h} = \frac{\mathcal{B}_l}{H}
\end{equation}
CAKE also uses \textbf{head attention loss} function as optimization objective but it also introduces  temporal information into score function of SnapKV. It adopts variance to represent the distribution shift of attention scores for the same token. Let $\gamma$  be a hyper-parameter to control the influence of temporal information, and $w$ as the sliding window size, CaKE score is:
\begin{align}
    &s_{l,h}[i]=\sum_{j=N-w}^N A_{l,h}^{j}[i] + 
    \gamma \text{VAR}([A_{l,h}^t[i]]^{t\in[i,N]}) \notag \\ 
    &\mathcal{I}_{l,h}=Select(s_{l,h},\mathcal{B}_{l,h})
\end{align}

\paragraph{AdaKV.} \cite{feng2024ada} The algorithm of AdaKV is based on other methods. It adopts \textbf{layer attention output loss} function but not conduct real training. Deriving the upper bound of output loss (as shown in Eq.~\ref{ada bound} where $C=max_{h\in[H]}\|{W_{l,h}^O}^TV_{l,h}^T\|_1$), AdaKV obtains the insight that allocating different budgets to heads of one layer based on the score function just considering about information within attention scores can preserve the performance of model further. 
\begin{equation}
    \|y_l-\hat{y}_l\|_1 \leq 2 C \sum_{h\in[H]}(\sum_{i\in[N]}A_{l,h}^N[i](1-\mathcal{I}_{l,h}[i]))
    \label{ada bound}
\end{equation}
We set $\hat{s}_l$ as the topk results of all $s_{l,h}, h\in[H]$, the budget of one head $h$ can be calculated by:
\begin{equation}
    \mathcal{B}_{l,h} = Num(\hat{s}_{l,h}), \hat{s}_l,\mathcal{I}_l = Select(s_{l,h},\mathcal{B}_{l,h})
\end{equation}
AdaKV combines this insight with SnapKV and PyramidKV for better results. So the score function of AdaKV is the same as Eq.~\ref{accumulated and pooling attention scores}. However, the bound of AdaKV ignores the influence of value information and just use the max information, which will make the bound too loose. Our framework about output loss is motivated by this research and we conduct some modification and further studies. For the details and how to derive upper bound of output loss, refer to Section~\ref{methodology}. 

\paragraph{DuoAttention.} \cite{xiao2025duoattention}
DuoAttention uses \textbf{layer attention output loss} function as optimization objective. Unlike H2O and TOVA, attention mask $\mathcal{I}$ of DuoAttention is constraint to a pattern combined with sink and recent tokens based on allocation budgets $\mathcal{B}$, which means score function $s$ id for tokens are not needed. Here sink tokens means several initial tokens in prompt defined by StreamingLLM \cite{xiao2024efficient}.
\begin{equation}
    \mathcal{I}_{l,h}[i] =
    \begin{cases} 
    1 & \text{if position } k \text{ is sink or recent} , k\in [N] \\
    0 & \text{otherwise, evict } K_{l,h}[k] \text{ and } V_{l,h}[k]
    \end{cases}  
\end{equation}
DuoAttention adopts real optimization method and needs training based on 2-norm of output loss function. The optimization result is to determine the allocation budgets $\mathcal{B}$. In detail, it determines which head was allocated with full budget and which head was allocated with a compressed budget. So besides $\mathcal{I}$ and $\mathcal{B}$, DuoAttention introduces a parameter $\alpha$ to be optimized and finally determines the different functions of heads, including Retrieval Heads \cite{wu2024retrievalheadmechanisticallyexplains} and Streaming Heads. We define $\hat{w}$ as the numbers of sink and recent tokens.
\begin{equation} 
    \mathcal{B}_{l,h} =
    \begin{cases} 
    n & \text{if head } h \text{ of layer } l \text{ is Retrieval Head} \\
    \hat{w} & \text{otherwise, Streaming Head} 
    \end{cases}  
\end{equation}

\paragraph{Dodo.} \cite{qin2024dodo}
Dodo uses \textbf{logit loss} function as optimization objective. But not adopting a predefined rule for attention mask $\mathcal{I}$, Dodo uses a score function $\phi$ implemented by LoRA \cite{hu2021lora} adapters to determine the attention mask for tokens, which is trained along with logits loss. Logits loss is defined by loss of future expected tokens which are not pratical. So Dodo converts the expected tokens into past tokens and the loss function can be formalized as:
\begin{align}
    &P(\mathcal{I},\mathcal{B}) = \sum_{i\in[N]}CE(p, \hat{p})^i 
\end{align}
The score function $\phi$ is trained via this loss function and finally determines which tokens will be preserved. The cache budget $\mathcal{B}$ for all heads and layers are the same. Besides, Dodo merges the information within tokens evicted into the preserved tokens similar to KV Cache merging methods.

 \paragraph{VATP.}\cite{guo-etal-2024-attention} The difference between LAVa and VATP is shown in Table~\ref{tab:summary full} and explained as follows: (1) VATP directly multiplies each token’s value norm with attention scores. In contrast, LAVa calculates the maximum value norm, which serves as scaling factors for heads; (2) VATP has fixed head and layer budgets, while LAVa is totally dynamic. The deeper difference, however, lies in how the two scores are developed. VATP comes with an intuition of "Value also matters" but lacks theoretical analysis. We independently derive from layer attention output with a complete reasoning process: starting from layer attention output, deriving the upper bound, getting an approximate score in greedy solution, smoothing it out based on multiple residual stream.

This reasoning is very important. As we start from the layer point of view, we can see that such scores can be used to compare entries across heads for layer-wise KV Cache eviction. And we argue that doing so could reduce the information loss at layer attention output. The reasoning process shows what approximation we make and gives room for future improvement. 

To validate our elaboration, we compares three configurations: (1) VATP integrated with SnapKV, (2) standard LAVa, and (3) LAVa without dynamic layer budgeting based on Mistral-7B-Instruct-v0.2 in LongBench. The results in Table~\ref{tab:vatp ab} demonstrate that while VATP shows improvement over baseline SnapKV, it consistently underperforms compared to both LAVa and LAVa (-layer dynamic). From the computational perspective, VATP incurs similar overhead to LAVa(refer to Appendix~\ref{details of experiments}, yet delivers suboptimal performance. This verifies our claim that intuition and a theoretical analysis help you get to a more optimal solution.

\begin{table}[t]
\centering
\begin{adjustbox}{max width=\textwidth}
\begin{tabular}{lcccc}
\toprule
 \textbf{Budgets} & 128 & 256 & 512 & 1024  \\ 
\midrule
\textbf{SnapKV} & 34.42 & 38.49 & 41.48 & 43.00 \\
\textbf{\ \ +VATP} & 35.34 & 39.41 & 41.93 & 43.32  \\
\textbf{LAVa} & \textbf{36.74} & \textbf{40.12} & \textbf{42.59} & \textbf{43.65}   \\
\textbf{\ \ -layer dynamic} & 36.20 & 39.77 & 42.11 & 43.35  \\
\bottomrule
\end{tabular}
\end{adjustbox}
\caption{Comparison between VATP and LAVa.}
\label{tab:vatp ab}
\end{table}

\section{Extension of LAVa: Layer-wise Cache Eviction with Dynamic Budget Allocation}
\label{sec:appendix}

\paragraph{Details of Lemma~\ref{thm:layer-output-loss}. }We define and derive the \textbf{Layer Attention Output Loss} in this lemma.
\label{proof of layer output loss}

\paragraph{Lemma 1.} Based on the $L_p$ norm, the layer attention output loss due to the attention mask $\mathcal{I}$ is measured for layer-$l$ at the current ($N$-th) decoding step as follows:
    \begin{align}\small
&\mathcal{P}(\textbf{x}^{1\ldots N}_1,\mathcal{I},\mathcal{B})=\|y_l^N-\hat{y}_l^N\|_p \\
    &= \left\|Cat_{h}\left[(A^N_{l,h}-\frac{A^N_{l,h}\odot \mathcal{I}_{l,h}}{\|A^N_{l,h}\odot\mathcal{I}_{l,h}\|_1})V_{l,h}\right]W_l^O\right\|_p \nonumber
    \label{output perturbation}
\end{align}
where $\odot$ indicates element-wise multiplication and $\hat{y}_l^N=Cat_h(\hat{A}_{l,h}^NV_{l,h})W_l^O$
\\
\noindent As we mentioned above:
\begin{align}
    y^N_l &= Cat_{h\in[H]}(A^N_{l,h}V_{l,h})W_l^O \notag\\
    \hat{y}_l^N&=Cat_{h\in[H}(\hat{A}_{l,h}^NV_{l,h})W_l^O \notag\\
\end{align}
And based on the definition of attention mask $\mathcal{I}$, the attention weights after eviction can be calculated as:
\begin{equation}
    \hat{A}^N_{l,h} = Softmax(\frac{-\inf\odot (\mathbf{1}-\mathcal{I}_{l,h})+Q^N_{l,h}{K_{l,h}^T}}{\sqrt{d_h}})
\end{equation}
Hence, Lemma~\ref{output perturbation} is equal to (Temporarily ignoring the superscript $N$):
\begin{equation}
    \hat{A}_{l,h} = \frac{A_{l,h}\odot \mathcal{I}_{l,h}}{\|A_{l,h}\odot\mathcal{I}_{l,h}\|_1}     
\end{equation}
This theorem has been proved by AdaKV \cite{feng2024ada}, so we will not elaborate further here.

\paragraph{Proof of Theorem~\ref{upper bound of output}. }Then we drive the \textbf{upper bound} of Layer Attention Output Loss and give this theorem. 
\label{proof of upper bound}
\paragraph{Theorem 1.} The $L_1$ norm of layer attention output loss can be bounded by:
\begin{dmath}
    \|y_l-\hat{y}_l\|_1 \\
     \leq 2\hat{C}\sum^{h\in[H]}\bar{V}_{l,h}(\sum^{k\in[N]}A_{l,h}^N[k](1-\mathcal{I}_{l,h}[k]))
\end{dmath}
 where $\bar{V}_{l,h} = max_{k\in[N]}\|V_{l,h}[k]\|_1$ and $\hat{C}=\|{{W_l^O}^T}\|_1$ is a constant, which is independent of any head or token within layer-$l$.
\begin{proof}
First we need to introduce a lemma:
\begin{lemma}
    Given a vector $x\in \mathbb{R}^{1\times m}$ and a matrix $W\in \mathbb{R}^{m\times n}$, we can get the relationship between matrix norm and vector norm:
    \begin{equation}
        \|xW\|_p \leq \|x\|_p\|W^T\|_p
    \end{equation}
    $\|xW\|_p$ and $\|x\|_p$ are vector p-norm, $\|W^T\|_p$ is matrix p-norm which is calculated by the largest sum of column absolute value.
    \label{norm inequality}
\end{lemma}
This lemma is derived from \citet{horn2012matrix}. Then we can obtain (Temporarily ignoring the superscript $N$):
\begin{dmath}
    \|y_l-\hat{y}_l\|_1 \\ \leq \|Cat_{h}[(A_{l,h}-\frac{A_{l,h}\odot \mathcal{I}_{l,h}}{\|A_{l,h}\odot\mathcal{I}_{l,h}\|_1})V_{l,h}]\|_1\|{W_l^O}^T\|_1
\end{dmath}
We set $\|{W_l^O}^T\|_1$ as $\hat{C}$ because it is the constant model parameter. Then we know that and set:
\begin{equation}
    G_{l,h}=(A_{l,h}-\frac{A_{l,h}\odot\mathcal{I}_{l,h}}{\|A_{l,h}\odot\mathcal{I}_{l,h}\|_1})V_{l,h} \in \mathbb{R}^{1\times d_h}
\end{equation}
Thus $\|Cat^{h\in[H]}[G_{l,h}]\|_1$ is the vector 1-norm of a vector $\in \mathbb{R}^{1\times (d_h*H)}$. According to the definition of vector 1-norm, we can transform cat operation to sum and continue derivation based on Theorem 2:
\begin{dmath}
    \|y_l-\hat{y}_l\|_1 \\ \leq \hat{C} \|Cat_{h\in[H]}[(A_{l,h}-\frac{A_{l,h}\odot \mathcal{I}_{l,h}}{\|A_{l,h}\odot\mathcal{I}_{l,h}\|_1})V_{l,h}]\|_1\\
     = \hat{C} \sum_{h\in[H]}\|(A_{l,h}-\frac{A_{l,h}\odot \mathcal{I}_{l,h}}{\|A_{l,h}\odot\mathcal{I}_{l,h}\|_1})V_{l,h}\|_1\\
     \leq \hat{C} \sum_{h\in[H]}(\|A_{l,h}-\frac{A_{l,h}\odot\mathcal{I}_{l,h}}{\|A_{l,h}\odot \mathcal{I}_{l,h}\|_1}\|_1\|V_{l,h}^T\|_1)
\end{dmath}
Next we will prove that $\|A_{l,h}-\frac{A_{l,h}\odot\mathcal{I}_{l,h}}{\|A_{l,h}\odot \mathcal{I}_{l,h}\|_1}\|_1 = 2\sum_{if\mathcal{I}_{l,h}[i]=0}^{i\in[N]}A_{l,h}[i]$. 

Let $\|A_{l,h}\odot\mathcal{I}_{l,h}\|_1=\sum_{i\in[N]}\mathcal{I}_{l,h}[i]A_{l,h}[i]=\sum^{i\in[N]}_{if\mathcal{I}_{l,h}[i]=1}A_{l,h}[i]$ as $F\in(0,1]$:

\begin{dmath}
    \|A_{l,h}-\frac{A_{l,h}\odot\mathcal{I}_{l,h}}{\|A_{l,h}\odot \mathcal{I}_{l,h}\|_1}\|_1  = \|\frac{F-\mathcal{I}_{l,h}}{F}\odot A_{l,h}\|_1\\
     = \sum_{i\in[N]}|\frac{(F-\mathcal{I}_{l,h}[i])A_{l,h}[i]}{F}| \\
     = \sum^{i\in[N]}_{if\mathcal{I}_{l,h}[i]=0}A_{l,h}[i]+\sum^{i\in[N]}_{if \mathcal{I}_{l,h}[i]=1}\frac{(1-F)A_{l,h}[i]}{F} \\
     = \sum^{i\in[N]}_{if\mathcal{I}_{l,h}[i]=0}A_{l,h}[i]+\frac{\sum_{if\mathcal{I}_{l,h}[i]=1}^{i\in[N]}A_{l,h}[i]}{F}\\-\sum_{if\mathcal{I}_{l,h}[i]=1}^{i\in[N]}A_{l,h}[i] \\
     = \sum^{i\in[N]}_{if\mathcal{I}_{l,h}[i]=0}A_{l,h}[i]+1-\sum_{if\mathcal{I}_{l,h}[i]=1}^{i\in[N]}A_{l,h}[i] \\
     = 2\sum_{if\mathcal{I}_{l,h}[i]=0}^{i\in[N]}A_{l,h}[i]
\end{dmath}
Then based on the definition of matrix 1-norm and $\|V_{l,h}^T\|_1\in\mathbb{R}^{d_h\times N}$, we can calculate this as the largest sum of row absolute value of $V_{l,h}\in\mathbb{R}^{N\times d_h}$, which is equals to the largest vector 1-norm of V value of previous tokens, formalized as:
\begin{equation}
    \bar{V}_{l,h} = \|V_{l,h}^T\|_1 = max_{k\in[N]}\|V_{l,h}[k]\|_1
\end{equation}
Now we can obtain:
\begin{dmath}
    \|y_l-\hat{y}_l\|_1 \\
     \leq 2\hat{C}\sum_{h\in[H]}(\sum^{i\in[N]}_{if\mathcal{I}_{l,h}[i]=0}A_{l,h}^N[i]\|V_{l,h}^T\|_1) \\
     =2\hat{C}\sum_{h\in[H]} (\sum_{i\in[N]}A_{l,h}^N[i]\bar{V}_{l,h}(1-\mathcal{I}_{l,h}[i]))
\end{dmath}
Here the proof is done.
\end{proof}


\paragraph{Potential Future Work.}
\label{future work}
 Building on our framework, multiple  research directions can be further explored.  One possible question is whether the \textit{Layer Output Loss}, which takes into account the FFN layer, should be considered. The interaction between the FFN layer and the layer attention output determines what information a layer writes to the residual stream \cite{ferrando-voita-2024-information}. In other words, certain tokens in past residual streams may play a crucial role in activating the layer’s knowledge within the FFN. Accounting for these interactions could reduce performance loss, yet the challenge lies in how to do so efficiently.

\noindent Another potential avenue is formulating the problem as an online reinforcement learning (RL) task, where the objective is to optimize the policy (i.e., the scoring function) to maximize the expected reward. Here, the expected reward can be cast as minimizing the expected loss in future residual streams, not just the past ones. This direction is potential for the cache-offload and retrieval problem, where we need to decide which parts of the cache to offload to CPU or retrieve from CPU while maintaining the communication cost.  

Additionally, this framework could be extended to model pruning, not just masking tokens but also selectively masking model parameters to minimize information flow while preserving efficiency. Also, the value of algorithms like LAVa and SnapKV, CAKE lies in their potential to serve as components of larger solutions tailored for long output and multi-turn contexts. These static methods  can be integrated with merging techniques , cache retrieval and offloading, quantization methods, or dynamic approaches. This is also a promising direction for our future work.

\section{Extension of Experiments}
\label{details of experiments}

\paragraph{Implementation Details.}
\label{ref:implementation}
For SnapKV and Ada-SnapKV, no additional hyperparameters are required. However, for PyramidKV, we must adjust the parameter \( \beta \) to control the shape of the cache budget pyramid. We set \( \beta \) to (5, 10, 20) and select the best-performing result, the same approach to Ada-PyramidKV.  For CAKE, three parameters require tuning: \( \gamma_1 \) and \( \gamma_2 \) for layer budget allocation, and \( \gamma_3 \) for the scoring function, as explained in Appendix~\ref{descriptions}. Based on recommendations from \cite{qin2025cake}, we set \( 1/\gamma_1 \) to (0.2, 0.3, 0.5, 1, 2), \( 1/\gamma_2 \) to (0.2, 0.3, 0.5, 1, 2), and \( \gamma_3 \) to (0, 5, 10, 200). We then evaluate different combinations and select the one that yields the best overall performance.

Pooling operators, such as max pooling or average pooling, can be applied to token score vectors to smooth score variations across adjacent tokens \cite{li2024snapkv,cai2024pyramidkv,qin2025cake}. This strategy is also employed in the implementation of LAVa and all the baselines. For pooling operation, for all methods, we adopt maxpool function and set kernel size as 7.

\paragraph{Results of LAVa in LongBench.}
\label{ref:results of our others longbench}
The results of Qwen2.5-7B-Instruct are listed in Table~\ref{our qwen2 longbench}. The results of Qwen2.5-14B-Instruct and Qwen2.5-32B-Instruct are in Table~\ref{others' longbench}. The results of Llama3-8B-Instruct are in Table~\ref{our llama3 longbench}. From all these results, we can obtain the similar conclusion like Mistral in main text. LAVa outperforms all baselines across different budgets, even in models with larger parameter size. 

\begin{table*}[htbp]
\centering
\begin{adjustbox}{max width=\textwidth}
\begin{tabular}{lccccccccccccccccccccc|c}
\toprule
\multirow{2}{*}{} & \multicolumn{4}{c}{\textbf{Single-Doc. QA}} & \multicolumn{4}{c}{\textbf{Multi-Doc. QA}} & \multicolumn{4}{c}{\textbf{Summarization}} & \multicolumn{4}{c}{\textbf{Few-shot Learning}} & \multicolumn{3}{c}{\textbf{Synthetic}} & \multicolumn{2}{c}{\textbf{Code}} & \multirow{1}{*}{}\\
\cmidrule(lr){2-5} \cmidrule(lr){6-9} \cmidrule(lr){10-13} \cmidrule(lr){14-17} \cmidrule(lr){18-20} \cmidrule(lr){21-22} 
& \rotatebox{-60}{NrtvQA} & \rotatebox{-60}{Qasper} & \rotatebox{-60}{MF-en} & \rotatebox{-60}{MF-zh} & \rotatebox{-60}{HotpotQA} & \rotatebox{-60}{2WikiMQA} & \rotatebox{-60}{Musique} & \rotatebox{-60}{Dureader} & \rotatebox{-60}{GovReport} & \rotatebox{-60}{QMSum} & \rotatebox{-60}{VCSUM} & \rotatebox{-60}{MultiNews} & \rotatebox{-60}{TREC} & \rotatebox{-60}{TriviaQA} & \rotatebox{-60}{SAMSum} & \rotatebox{-60}{LSHT} &  \rotatebox{-60}{PCount} & \rotatebox{-60}{PR-en} & \rotatebox{-60}{PR-zh} & \rotatebox{-60}{Lcc} & \rotatebox{-60}{RepoBench-P} & \rotatebox{-60}{Avg}\\
\midrule
\textbf{Full Cache} & 29.05 & 43.34 & 52.52 & 62.27 & 57.59 & 47.05 & 30.24 & 29.25 & 31.78 & 23.64 & 15.96 & 23.96 & 72.50 & 88.82 & 45.61 & 42.75 & 8.50 & 100.00 & 96.50 & 59.61 & 67.12 & 48.96\\

\midrule
\multicolumn{23}{c}{\textbf{$\mathbb{B}=128\mathrm{HL}$}}\\
\textbf{PyramidKV} & 21.96 & 26.41 & 42.53 & 52.77 & 49.33 & 42.17 & 23.48 & 17.88 & 16.80 & 19.29 & 11.24 & 14.30 & \ul{42.50} & 83.78 & 41.15 & 22.39 & \ul{8.50} & 95.50 & 63.50 & 48.53 & 51.39 & 37.88\\
\textbf{SnapKV} & \textbf{25.24} & 27.66 & 43.90 & 53.53 & 51.00 & 42.12 & 24.59 & 18.56 & 18.04 & 19.85 & 11.32 & 15.55 & 41.00 & 83.18 & 40.68 & \ul{24.88} & \textbf{9.00} & \textbf{98.00} & 81.50 & 49.44 & 52.58 & 39.60\\
\textbf{Ada-PyramidKV} & 23.08 & 27.53 & 42.07 & 53.17 & 50.73 & 42.03 & 23.31 & 18.03 & 17.48 & 19.65 & 11.21 & 14.71 & \ul{42.50} & 83.90 & 41.25 & 22.81 & \textbf{9.00} & 94.00 & 76.00 & 49.17 & 52.69 & 38.78\\
\textbf{Ada-SnapKV} & \ul{25.20} & 28.45 & 45.00 & 54.37 & \ul{51.08} & \textbf{44.02} & 24.66 & 18.81 & \textbf{18.26} & 20.09 & 11.50 & \ul{16.25} & \ul{42.50} & 84.06 & 41.00 & 22.49 & \textbf{9.00} & \ul{96.50} & \textbf{87.50} & 49.92 & 54.32 & 40.24\\
\textbf{CAKE} & 24.43 & \textbf{30.15} & \ul{45.03} & \ul{54.86} & 50.65 & 42.41 & \textbf{25.91} & \ul{18.89} & 18.21 & \textbf{20.66} & \ul{11.60} & 15.84 & 42.00 & \ul{84.54} & \ul{41.95} & \textbf{26.24} & \ul{8.50} & 95.50 & 81.50 & \ul{51.60} & \ul{55.09} & \ul{40.26}\\
\textbf{LAVa (Ours)} & 23.29 & \ul{28.87} & \textbf{46.80} & \textbf{56.10} & \textbf{52.65} & \ul{42.96} & \ul{25.09} & \textbf{19.25} & \ul{18.24} & \ul{20.52} & \textbf{11.80} & \textbf{16.28} & \textbf{43.00} & \textbf{84.56} & \textbf{42.18} & 23.95 & \ul{8.50} & 96.00 & \ul{85.00} & \textbf{53.45} & \textbf{56.07} & \textbf{40.69}\\
\midrule
\multicolumn{23}{c}{\textbf{$\mathbb{B}=256\mathrm{HL}$}}\\
\textbf{PyramidKV} & 24.82 & 31.13 & 46.92 & 56.06 & 53.07 & 42.31 & 25.06 & 19.54 & 19.27 & 20.47 & 12.01 & 16.55 & 50.00 & 84.88 & 42.04 & 25.39 & \textbf{8.50} & 96.00 & 85.50 & 52.03 & 55.82 & 41.30\\
\textbf{SnapKV} & \ul{26.61} & 23.77 & 49.15 & \ul{58.37} & \textbf{56.03} & \ul{44.18} & 25.68 & \textbf{20.96} & 20.84 & 20.99 & 12.19 & 18.52 & 48.50 & \ul{86.31} & 43.06 & 29.89 & \textbf{8.50} & \ul{97.50} & \textbf{95.00} & 54.26 & 59.42 & 43.32\\
\textbf{Ada-PyramidKV} & 25.97 & 31.01 & 47.31 & 56.43 & 54.17 & 43.03 & 25.23 & 19.41 & 19.60 & 21.09 & 11.87 & 17.07 & \textbf{54.50} & 86.04 & 42.69 & 27.28 & \textbf{8.50} & 97.00 & 90.00 & 52.78 & 56.55 & 42.26\\
\textbf{Ada-SnapKV} & 26.52 & \ul{34.50} & \textbf{50.01} & 58.28 & \ul{55.61} & 43.60 & 26.14 & \ul{20.89} & \textbf{21.30} & 20.94 & \ul{12.51} & \ul{18.59} & \ul{52.50} & 85.50 & 42.97 & 28.43 & \textbf{8.50} & \textbf{98.00} & 93.50 & 53.94 & 59.30 & 43.41\\
\textbf{CAKE} & 26.59 & 33.95 & \ul{49.80} & 58.25 & 54.89 & \textbf{44.42} & \ul{26.47} & 20.35 & \ul{21.23} & \textbf{21.94} & 12.35 & 18.53 & 47.50 & 85.41 & \textbf{43.51} & \textbf{32.33} & \textbf{8.50} & \ul{97.50} & \ul{94.00} & \ul{55.56} & \ul{61.13} & \ul{43.53}\\
\textbf{LAVa (Ours)} & \textbf{27.04} & \textbf{35.19} & 49.36 & \textbf{59.74} & 55.35 & 44.13 & \textbf{27.25} & 20.88 & 21.15 & \ul{21.51} & \textbf{12.77} & \textbf{18.96} & 49.00 & \textbf{86.73} & \ul{43.42} & \ul{30.35} & \textbf{8.50} & \textbf{98.00} & 93.00 & \textbf{56.19} & \textbf{62.19} & \textbf{43.84}\\
\midrule
\multicolumn{23}{c}{\textbf{$\mathbb{B}=512\mathrm{HL}$}}\\
\textbf{PyramidKV} & 28.02 & 35.74 & \textbf{50.84} & 58.11 & 55.26 & 44.72 & 25.85 & 20.94 & 21.83 & 21.34 & 12.33 & 18.95 & \ul{59.50} & 86.13 & 43.04 & 32.83 & \textbf{8.50} & 99.00 & \textbf{96.00} & 55.65 & 59.42 & 44.48\\
\textbf{SnapKV} & \textbf{28.27} & 28.22 & \ul{50.69} & \ul{60.27} & \textbf{56.18} & 44.69 & 27.28 & 21.98 & 23.79 & 21.89 & \textbf{13.20} & 20.64 & 59.50 & 84.10 & 43.68 & 35.52 & \textbf{8.50} & \textbf{100.00} & 94.00 & 56.66 & 62.69 & 45.32\\
\textbf{Ada-PyramidKV} & 27.31 & 37.36 & 49.62 & 58.57 & 55.40 & 44.66 & 26.74 & 21.35 & 22.39 & 21.12 & 12.42 & 19.32 & \textbf{62.00} & \textbf{86.29} & 43.78 & 33.33 & \textbf{8.50} & 99.00 & \ul{95.50} & 55.78 & 60.99 & 44.83\\
\textbf{Ada-SnapKV} & 28.03 & 38.51 & 50.06 & \textbf{60.54} & 55.50 & 45.06 & \textbf{28.81} & 22.04 & \textbf{23.98} & \ul{22.49} & \ul{13.05} & \ul{20.80} & \textbf{62.00} & 85.83 & 44.37 & 37.10 & \textbf{8.50} & \textbf{100.00} & 94.00 & 56.44 & 62.71 & \ul{45.71}\\
\textbf{CAKE} & \ul{28.17} & \textbf{39.09} & 50.22 & 60.00 & 54.89 & \ul{45.21} & 26.31 & \ul{22.20} & 23.65 & 21.98 & 13.04 & 20.57 & 57.50 & 85.60 & \ul{44.61} & \ul{37.23} & \textbf{8.50} & \ul{99.50} & 94.00 & \textbf{58.27} & \ul{63.95} & 45.45\\
\textbf{LAVa (Ours)} & 27.21 & \ul{39.08} & 50.47 & 60.09 & \ul{55.63} & \textbf{45.25} & \ul{27.75} & \textbf{22.91} & \ul{23.83} & 
\textbf{22.81} & \ul{13.05} & \textbf{20.84} & 58.50 & \ul{86.15} & \textbf{45.02} & \textbf{37.43} & \textbf{8.50} & \textbf{100.00} & 93.50 & \ul{58.02} & \textbf{64.57} & \textbf{45.74}\\
\midrule
\multicolumn{23}{c}{\textbf{$\mathbb{B}=1024\mathrm{HL}$}}\\
\textbf{PyramidKV} & 28.06 & 40.11 & 51.83 & 60.22 & \textbf{57.55} & 45.38 & 29.31 & 22.42 & 24.35 & 22.04 & 13.12 & 21.12 & 68.00 & 85.27 & 44.18 & 36.99 & \textbf{8.50} & \textbf{100.00} & \textbf{96.50} & 58.29 & 62.56 & 46.47\\
\textbf{SnapKV} & 29.01 & \ul{42.02} & \ul{51.86} & \textbf{61.22} & 56.82 & 45.04 & 28.95 & \ul{23.97} & \ul{26.26} & 22.76 & 13.66 & \textbf{22.50} & 68.50 & 86.85 & \textbf{45.52} & \textbf{42.50} & \textbf{8.50} & \textbf{100.00} & \textbf{96.50} & 57.94 & 65.59 & 47.43\\
\textbf{Ada-PyramidKV} & 28.52 & 40.50 & \textbf{51.87} & 60.27 & 56.42 & \textbf{45.80} & 29.18 & 23.01 & 24.45 & 22.10 & 13.31 & 21.25 & \ul{69.00} & 86.41 & 45.10 & 37.79 & \textbf{8.50} & \textbf{100.00} & \textbf{96.50} & 57.16 & 63.31 & 46.69\\
\textbf{Ada-SnapKV} & 29.61 & \textbf{42.30} & 51.79 & 60.29 & 56.38 & \ul{45.75} & 29.30 & 23.64 & 26.21 & 22.80 & \textbf{13.85} & 22.39 & \ul{69.00} & \textbf{88.09} & 45.36 & 41.75 & \textbf{8.50} & \textbf{100.00} & \ul{96.00} & 58.15 & 65.77 & \ul{47.47}\\
\textbf{CAKE} & \ul{29.70} & 41.08 & 51.85 & 60.64 & \ul{57.34} & 45.02 & \textbf{30.48} & 23.82 & 25.92 & \textbf{22.95} & 13.69 & \ul{22.45} & 67.50 & 86.63 & 45.22 & \ul{42.00} & \textbf{8.50} & \textbf{100.00} & \textbf{96.50} & \ul{59.49} & \ul{65.99} & \ul{47.47}\\
\textbf{LAVa (Ours)} & \textbf{29.79} & 41.68 & 51.84 & \ul{60.79} & 57.04 & 45.27 & \ul{30.01} & \textbf{23.99} & \textbf{26.36} & \ul{22.90} & \ul{13.81} & 22.42 & \textbf{69.50} & \ul{87.42} & \ul{45.46} & 41.00 & \textbf{8.50} & \textbf{100.00} & \textbf{96.50} & \textbf{59.97} & \textbf{66.24} & \textbf{47.64}\\

\bottomrule
\end{tabular}
\end{adjustbox}
\caption{Final comparison based on Qwen2.5-7B-Instruct among 21 datasets of LongBench. (Note: The best result is highlighted in \textbf{bold}, and the second is in \ul{underline}. )}
\label{our qwen2 longbench}
\end{table*}

\begin{table*}[htbp]
\centering
\begin{adjustbox}{max width=\textwidth}
\begin{tabular}{lccccccccccccccccccccc|c}
\toprule
\multirow{2}{*}{} & \multicolumn{4}{c}{\textbf{Single-Doc. QA}} & \multicolumn{4}{c}{\textbf{Multi-Doc. QA}} & \multicolumn{4}{c}{\textbf{Summarization}} & \multicolumn{4}{c}{\textbf{Few-shot Learning}} & \multicolumn{3}{c}{\textbf{Synthetic}} & \multicolumn{2}{c}{\textbf{Code}} & \multirow{1}{*}{}\\
\cmidrule(lr){2-5} \cmidrule(lr){6-9} \cmidrule(lr){10-13} \cmidrule(lr){14-17} \cmidrule(lr){18-20} \cmidrule(lr){21-22} 
& \rotatebox{-60}{NrtvQA} & \rotatebox{-60}{Qasper} & \rotatebox{-60}{MF-en} & \rotatebox{-60}{MF-zh} & \rotatebox{-60}{HotpotQA} & \rotatebox{-60}{2WikiMQA} & \rotatebox{-60}{Musique} & \rotatebox{-60}{Dureader} & \rotatebox{-60}{GovReport} & \rotatebox{-60}{QMSum} & \rotatebox{-60}{VCSUM} & \rotatebox{-60}{MultiNews} & \rotatebox{-60}{TREC} & \rotatebox{-60}{TriviaQA} & \rotatebox{-60}{SAMSum} & \rotatebox{-60}{LSHT} &  \rotatebox{-60}{PCount} & \rotatebox{-60}{PR-en} & \rotatebox{-60}{PR-zh} & \rotatebox{-60}{Lcc} & \rotatebox{-60}{RepoBench-P} & \rotatebox{-60}{Avg}\\
\midrule
\multicolumn{23}{c}{\textbf{Qwen2.5-14B-Instruct}}\\
\midrule
\textbf{Full Cache} & 29.33 & 45.19 & 53.59 & 62.79 & 62.59 & 57.69 & 38.47 & 29.87 & 29.74 & 23.53 & 14.75 & 21.90 & 77.50 & 90.23 & 47.27 & 50.00 & 9.23 & 98.67 & 98.25 & 62.60 & 51.13 & 50.21\\
\midrule
\multicolumn{23}{c}{\textbf{Qwen2.5-14B-Instruct, B=128h}}\\
\textbf{PyramidKV} & 19.67 & 22.26 & 39.57 & 50.04 & 50.75 & 49.47 & 30.31 & 16.67 & 16.10 & 19.43 & 10.53 & 13.51 & 42.00 & 82.29 & 40.90 & 27.00 & \ul{12.12} & 82.50 & 56.67 & 54.52 & 41.38 & 37.03\\
\textbf{SnapKV} & 21.04 & 25.50 & 42.11 & 49.89 & 54.31 & 51.87 & \ul{33.60} & 17.78 & 17.12 & 19.95 & 10.75 & 14.53 & 43.50 & 85.95 & 41.81 & 26.75 & 10.50 & 89.58 & 65.00 & 55.42 & 43.42 & 39.07\\
\textbf{Ada-PyramidKV} & 20.85 & 24.83 & 40.88 & 51.78 & 54.65 & 52.34 & 29.78 & 16.83 & 16.67 & 19.59 & 10.32 & 13.90 & \textbf{46.50} & 80.76 & 40.58 & 25.75 & 11.18 & 87.75 & 63.75 & 53.72 & 43.49 & 37.90\\
\textbf{Ada-SnapKV} & 22.16 & 25.58 & \ul{42.80} & \ul{52.22} & \ul{55.10} & 53.21 & 33.50 & \ul{17.98} & \ul{17.69} & \ul{20.25} & \ul{10.86} & 14.81 & 45.50 & 85.62 & \ul{42.49} & 27.00 & 9.05 & \ul{91.33} & 68.17 & \textbf{56.26} & 43.39 & 39.76\\
\textbf{CAKE} & \ul{22.20} & \ul{26.13} & 42.10 & 50.83 & 54.75 & \ul{53.25} & 31.77 & 17.73 & 17.56 & 19.98 & 10.84 & \textbf{15.44} & 44.00 & \textbf{87.51} & \textbf{42.65} & \textbf{28.50} & \textbf{13.96} & 86.50 & \textbf{78.83} & 54.92 & \ul{43.90} & \ul{40.16}\\
\textbf{LAVa (Ours)} & \textbf{22.24} & \textbf{26.52} & \textbf{43.09} & \textbf{52.39} & \textbf{55.97} & \textbf{53.43} & \textbf{33.68} & \textbf{18.23} & \textbf{17.94} & \textbf{20.57} & \textbf{10.98} & \ul{15.10} & \ul{46.00} & \ul{86.79} & 42.20 & \ul{27.17} & 10.53 & \textbf{92.00} & \ul{73.00} & \ul{55.74} & \textbf{44.63} & \textbf{40.39}\\
\midrule
\multicolumn{23}{c}{\textbf{Qwen2.5-14B-Instruct, B=512h}}\\
\textbf{PyramidKV} & 26.18 & 38.19 & 48.71 & 59.81 & \textbf{60.74} & 55.26 & 36.82 & 20.55 & 21.21 & 21.27 & 11.86 & 18.43 & 68.50 & 89.21 & 45.38 & 44.25 & \ul{8.59} & \textbf{98.33} & 96.75 & 59.71 & 48.71 & 46.59\\
\textbf{SnapKV} & \textbf{26.99} & 39.34 & 48.84 & 59.34 & 60.20 & 54.86 & 37.47 & \ul{21.43} & 22.25 & 21.95 & 11.93 & 19.34 & 66.50 & 88.78 & 45.95 & 45.25 & 8.22 & \ul{98.25} & \textbf{98.58} & 61.12 & 49.42 & 46.95\\
\textbf{Ada-PyramidKV} & \ul{26.78} & 40.25 & \textbf{49.71} & \ul{60.40} & \ul{60.64} & \textbf{55.69} & 37.72 & 20.75 & 21.49 & 21.54 & 11.67 & 18.60 & \textbf{70.00} & 88.59 & 45.70 & 44.50 & \textbf{8.77} & \textbf{98.33} & 96.75 & 60.23 & 48.85 & 47.00\\
\textbf{Ada-SnapKV} & 26.03 & \textbf{41.56} & \ul{49.42} & \textbf{60.88} & 59.99 & \ul{55.63} & \ul{38.34} & 21.33 & 22.49 & \ul{22.09} & \ul{11.96} & 19.32 & \ul{69.50} & \ul{89.01} & \ul{46.35} & \textbf{46.75} & 7.72 & 98.17 & \ul{98.50} & \textbf{62.21} & \ul{49.92} & \textbf{47.48}\\
\textbf{CAKE} & 25.39 & 39.92 & 48.62 & 60.30 & 60.42 & 55.19 & \textbf{38.37} & 21.40 & \ul{22.56} & 21.72 & \textbf{12.31} & \textbf{19.57} & \textbf{70.00} & \textbf{89.03} & 46.19 & \ul{46.25} & 6.68 & 98.17 & 98.25 & 60.90 & 49.31 & 47.17\\
\textbf{LAVa (Ours)} & 26.23 & \ul{40.65} & 48.93 & 59.45 & 60.34 & 55.36 & 37.50 & \textbf{21.53} & \textbf{22.57} & \textbf{22.13} & 11.91 & \ul{19.48} & 67.00 & 88.68 & \textbf{46.50} & \textbf{46.75} & 7.98 & 97.75 & 97.75 & \ul{61.85} & \textbf{50.38} & \ul{47.18}\\
\midrule
\multicolumn{23}{c}{\textbf{Qwen2.5-32B-Instruct}}\\
\midrule
\textbf{Full Cache} & \multicolumn{20}{c}{\textbf{OOM}}\\
\midrule
\multicolumn{23}{c}{\textbf{Qwen2.5-32B-Instruct, B=128h}}\\
\textbf{PyramidKV} & 21.32 & 27.86 & 43.55 & 56.05 & 55.74 & 53.85 & 32.25 & 16.74 & 17.08 & 18.88 & 10.71 & 15.76 & 48.00 & 54.41 & \textbf{40.69} & 29.50 & 11.17 & 94.00 & 73.09 & 48.04 & 35.36 & 38.29\\
\textbf{SnapKV} & 21.72 & 28.31 & 42.83 & 56.03 & 54.43 & 55.52 & 30.78 & 16.94 & 16.92 & 19.04 & 10.53 & 15.69 & 48.50 & \ul{58.30} & 39.64 & 27.50 & \ul{12.00} & 93.75 & 74.37 & 47.15 & 35.82 & 38.37\\
\textbf{Ada-PyramidKV} & 21.19 & \ul{29.67} & \textbf{45.61} & \textbf{58.04} & \textbf{57.30} & 55.65 & \ul{32.96} & 17.45 & 17.37 & 19.30 & \ul{10.89} & 16.02 & \textbf{51.50} & 56.24 & 40.24 & \ul{30.25} & \ul{12.00} & \ul{97.00} & 82.67 & \textbf{48.14} & 35.94 & \ul{39.78}\\
\textbf{Ada-SnapKV} & \ul{21.79} & 28.64 & 45.49 & 56.56 & \ul{57.12} & \ul{56.14} & 32.54 & \textbf{17.66} & 17.63 & 19.31 & 10.66 & 16.12 & \ul{49.50} & \textbf{60.07} & 40.03 & 27.50 & \ul{12.00} & 96.04 & \textbf{85.13} & 47.96 & \ul{36.29} & 39.72 \\
\textbf{CAKE} & 21.28 & 28.40 & 43.30 & 55.71 & 55.93 & 54.89 & 32.86 & 17.04 & 17.00 & \ul{19.44} & 10.50 & \ul{16.18} & 46.50 & 56.35 & \ul{40.38} & \textbf{31.88} & \textbf{12.50} & 94.79 & 82.92 & 46.63 & 36.05 & 39.07\\
\textbf{LAVa (Ours)} & \textbf{22.29} & \textbf{30.12} & \ul{45.50} & \ul{57.06} & 56.59 & \textbf{58.51} & \textbf{33.72} & \ul{17.50} & 17.42 & \textbf{19.97} & \textbf{11.09} & \textbf{16.29} & 48.50 & 57.21 & 40.23 & 28.17 & 10.00 & \textbf{97.42} & \ul{84.09} & \ul{48.12} & \textbf{36.68} & \textbf{39.83} \\
\midrule
\multicolumn{23}{c}{\textbf{Qwen2.5-32B-Instruct, B=512h}}\\
\textbf{PyramidKV} & 26.00 & 37.40 & 48.67 & 61.17 & 60.60 & 60.44 & 34.75 & 19.37 & 20.84 & 20.61 & 11.64 & 18.48 & 66.00 & 55.11 & 42.71 & 39.00 & \textbf{11.56} & \ul{99.75} & 98.54 & 50.28 & 38.12 & 43.86\\
\textbf{SnapKV} & 25.71 & 40.23 & 48.81 & 62.94 & 61.16 & 60.60 & 34.85 & \ul{20.64} & 22.69 & 21.27 & 11.61 & 20.04 & 66.50 & \ul{77.77} & \ul{44.01} & 41.86 & 11.19 & \textbf{100.00} & \ul{99.03} & 52.20 & \textbf{39.15} & 45.82\\
\textbf{Ada-PyramidKV} & 26.41 & 38.97 & \ul{50.14} & 61.50 & 61.50 & \textbf{61.86} & \textbf{37.55} & 19.67 & 21.49 & 20.71 & 11.23 & 18.68 & 67.50 & 60.81 & 43.40 & 39.75 & 11.08 & \ul{99.75} & \textbf{99.62} & 50.60 & 38.27 & 44.79\\
\textbf{Ada-SnapKV} & \textbf{27.51} & 39.44 & 49.21 & \ul{63.09} & \ul{61.70} & \ul{61.60} & 37.23 & 20.35 & 22.69 & \ul{21.72} & \ul{11.74} & \textbf{20.45} & \textbf{69.00} & \textbf{77.87} & \textbf{44.19} & 42.04 & \textbf{11.56} & \textbf{100.00} & 98.24 & 52.22 & \ul{39.14} & \ul{46.24}\\
\textbf{CAKE} & 25.32 & \ul{40.24} & 49.66 & \textbf{63.28} & 59.75 & 61.42 & 37.11 & 20.44 & \ul{22.73} & 21.22 & 11.67 & 20.28 & 66.50 & 77.31 & 43.92 & \textbf{44.58} & 11.19 & \textbf{100.00} & 98.78 & \textbf{52.36} & 38.99 & 46.04\\
\textbf{LAVa (Ours)} & \ul{26.56} & \textbf{41.18} & \textbf{50.80} & 62.49 & \textbf{61.90} & 60.83 & \ul{37.25} & \textbf{21.44} & \textbf{23.16} & \textbf{22.02} & \textbf{11.86} & \ul{20.30} & \ul{68.50} & 77.69 & 43.97 & \ul{42.23} & \ul{11.50} & \textbf{100.00} & 98.53 & \ul{52.24} & 38.86 & \textbf{46.35}\\

\bottomrule
\end{tabular}
\end{adjustbox}
\caption{Final comparison based on Qwen2.5-14B-Instruct and Qwen2.5-32B-Instruct among 21 datasets of LongBench. (Note: The best result is highlighted in \textbf{bold}, and the second is in \ul{underline}.)}
\label{others' longbench}
\end{table*}

\begin{table*}[htbp]
\centering
\begin{adjustbox}{max width=\textwidth}
\begin{tabular}{lccccccccccccccccccccc|c}
\toprule
\multirow{2}{*}{} & \multicolumn{4}{c}{\textbf{Single-Doc. QA}} & \multicolumn{4}{c}{\textbf{Multi-Doc. QA}} & \multicolumn{4}{c}{\textbf{Summarization}} & \multicolumn{4}{c}{\textbf{Few-shot Learning}} & \multicolumn{3}{c}{\textbf{Synthetic}} & \multicolumn{2}{c}{\textbf{Code}} & \multirow{1}{*}{}\\
\cmidrule(lr){2-5} \cmidrule(lr){6-9} \cmidrule(lr){10-13} \cmidrule(lr){14-17} \cmidrule(lr){18-20} \cmidrule(lr){21-22} 
& \rotatebox{-60}{NrtvQA} & \rotatebox{-60}{Qasper} & \rotatebox{-60}{MF-en} & \rotatebox{-60}{MF-zh} & \rotatebox{-60}{HotpotQA} & \rotatebox{-60}{2WikiMQA} & \rotatebox{-60}{Musique} & \rotatebox{-60}{Dureader} & \rotatebox{-60}{GovReport} & \rotatebox{-60}{QMSum} & \rotatebox{-60}{VCSUM} & \rotatebox{-60}{MultiNews} & \rotatebox{-60}{TREC} & \rotatebox{-60}{TriviaQA} & \rotatebox{-60}{SAMSum} & \rotatebox{-60}{LSHT} &  \rotatebox{-60}{PCount} & \rotatebox{-60}{PR-en} & \rotatebox{-60}{PR-zh} & \rotatebox{-60}{Lcc} & \rotatebox{-60}{RepoBench-P} & \rotatebox{-60}{Avg}\\
\midrule
\textbf{Full Cache} & 21.23 & 43.25 & 44.68 & 57.47 & 48.12 & 38.55 & 24.72 & 27.46 & 29.97 & 22.19 & 0.18 & 27.32 & 73.00 & 90.50 & 42.04 & 23.50 & 7.00 & 70.50 & 93.00 & 60.76 & 48.85 & 44.71\\

\midrule
\multicolumn{23}{c}{\textbf{$\mathbb{B}=128\mathrm{HL}$}}\\
\textbf{PyramidKV} & 18.36 & 30.37 & 42.49 & 49.83 & \textbf{47.59} & 35.11 & 23.20 & 18.65 & 18.62 & 20.95 & 0.07 & 20.27 & 49.50 & 88.67 & \ul{38.52} & \textbf{21.00} & \ul{3.75} & \ul{68.50} & 91.00 & 58.54 & 50.29 & 39.76\\
\textbf{SnapKV} & \textbf{18.88} & 30.10 & 41.33 & \textbf{50.68} & \ul{47.16} & 34.47 & 23.91 & 18.73 & 18.43 & 20.92 & 0.08 & 19.63 & 46.50 & 87.98 & 38.25 & 20.00 & 3.50 & 68.00 & 89.50 & 59.16 & 52.97 & 39.51\\
\textbf{Ada-PyramidKV} & 16.62 & \textbf{32.27} & \textbf{43.41} & \ul{50.49} & 47.07 & \textbf{35.70} & \textbf{24.09} & \textbf{19.27} & \textbf{19.63} & 21.14 & 0.06 & \textbf{20.96} & \textbf{58.50} & 89.10 & 38.41 & \textbf{21.00} & \textbf{4.00} & \ul{68.50} & \textbf{93.50} & 59.27 & 51.72 & \textbf{40.73}\\
\textbf{Ada-SnapKV} & 17.78 & 30.31 & 42.60 & 49.81 & 46.94 & 34.97 & 23.72 & \ul{19.16} & \ul{19.21} & \textbf{21.28} & 0.09 & 20.37 & \ul{55.50} & \textbf{89.68} & \textbf{38.96} & \textbf{21.00} & \textbf{4.00} & \textbf{69.00} & \textbf{93.50} & \textbf{60.32} & \ul{53.52} & \ul{40.58}\\
\textbf{CAKE} & \ul{18.40} & 30.92 & 42.41 & 49.50 & 47.04 & 35.08 & \ul{24.03} & 18.77 & 18.77 & 20.95 & 0.07 & 19.74 & 47.00 & \ul{89.34} & 38.03 & \ul{20.75} & \textbf{4.00} & \ul{68.50} & \ul{92.00} & 58.97 & 53.33 & 39.88\\
\textbf{LAVa (Ours)} & 17.24 & \ul{31.54} & \ul{43.26} & 49.96 & 46.69 & \ul{35.17} & \textbf{24.09} & 18.99 & 19.02 & \ul{21.19} & 0.12 & \ul{20.52} & 52.50 & 89.22 & 38.45 & 20.00 & \textbf{4.00} & \textbf{69.00} & \textbf{93.50} & \ul{59.74} & \textbf{53.58} & 40.38\\
\midrule
\multicolumn{23}{c}{\textbf{$\mathbb{B}=256\mathrm{HL}$}}\\
\textbf{PyramidKV} & 18.34 & 36.26 & 43.69 & 52.91 & 47.19 & 36.44 & \textbf{25.03} & 20.28 & 20.59 & \textbf{21.57} & 0.14 & 22.50 & 61.50 & 88.77 & \ul{39.26} & 21.75 & \ul{5.00} & \textbf{70.00} & 93.00 & 60.36 & 49.74 & 41.71\\
\textbf{SnapKV} & \ul{19.37} & 35.25 & 42.82 & 52.53 & 46.57 & \textbf{36.88} & 24.68 & 20.10 & 20.75 & 21.15 & 0.16 & 22.39 & 61.00 & 89.82 & 38.94 & 22.00 & \ul{5.00} & \textbf{70.00} & \textbf{94.00} & 61.16 & 52.57 & 41.84\\
\textbf{Ada-PyramidKV} & 18.71 & \textbf{37.29} & 44.59 & \textbf{53.63} & 47.26 & 36.39 & 24.33 & \textbf{20.92} & \ul{21.06} & 21.32 & 0.14 & \textbf{23.22} & \textbf{65.50} & 89.81 & 39.07 & \textbf{22.75} & 4.50 & \ul{69.50} & 92.50 & 61.00 & 50.27 & 42.18\\
\textbf{Ada-SnapKV} & 18.65 & 36.66 & \ul{45.14} & 52.62 & \ul{47.27} & 36.29 & \ul{24.84} & 19.91 & \textbf{21.41} & 21.14 & 0.15 & 22.70 & \textbf{65.50} & 89.72 & 39.15 & \ul{22.25} & \textbf{5.50} & \ul{69.50} & 93.00 & \ul{62.11} & 51.91 & \ul{42.26}\\
\textbf{CAKE} & \textbf{19.41} & 35.81 & 43.14 & 51.81 & 47.19 & \ul{36.76} & 24.76 & 20.30 & 20.81 & 21.02 & 0.12 & 22.50 & 60.50 & \ul{89.94} & 39.09 & 21.75 & \ul{5.00} & \ul{69.50} & \ul{93.50} & 61.45 & \ul{52.65} & 41.84\\
\textbf{LAVa (Ours)} & 19.16 & \ul{37.04} & \textbf{45.15} & \ul{53.58} & \textbf{47.87} & 36.57 & 24.61 & \ul{20.84} & \textbf{21.41} & \ul{21.41} & 0.15 & \ul{23.14} & \ul{65.00} & \textbf{90.36} & \textbf{39.93} & 22.00 & \textbf{5.50} & \textbf{70.00} & 93.00 & \textbf{62.70} & \textbf{53.69} & \textbf{42.65}\\
\midrule
\multicolumn{23}{c}{\textbf{$\mathbb{B}=512\mathrm{HL}$}}\\
\textbf{PyramidKV} & 19.63 & 40.70 & 43.99 & 55.36 & 47.30 & 37.81 & \ul{24.90} & \textbf{22.43} & 22.78 & 21.44 & 0.16 & 24.38 & 68.50 & \textbf{90.23} & 40.18 & \textbf{23.75} & \textbf{7.00} & \textbf{70.50} & \ul{93.00} & 61.83 & 49.51 & 43.26\\
\textbf{SnapKV} & \ul{19.67} & 40.65 & 45.65 & 54.68 & \ul{47.57} & 37.38 & 24.37 & 21.39 & 22.65 & \ul{21.78} & 0.17 & 24.44 & 68.00 & 90.19 & 40.32 & 23.25 & \textbf{7.00} & \ul{70.00} & \ul{93.00} & \ul{62.59} & 50.51 & 43.25\\
\textbf{Ada-PyramidKV} & 19.46 & 41.42 & \textbf{46.19} & \textbf{55.86} & \textbf{47.72} & \ul{37.90} & 24.54 & 21.98 & 22.91 & \textbf{21.84} & 0.12 & \ul{24.85} & \ul{69.50} & \textbf{90.23} & \ul{40.60} & \ul{23.50} & \ul{6.50} & \textbf{70.50} & \textbf{93.50} & 61.64 & 49.77 & 43.52\\
\textbf{Ada-SnapKV} & 19.38 & 
 41.03 & 45.05 & \ul{55.53} & 47.30 & 37.75 & 24.61 & 21.73 & \ul{23.10} & 21.59 & 0.15 & 24.70 & \textbf{70.50} & \textbf{90.23} & 40.49 & \textbf{23.75} & \ul{6.50} & \textbf{70.50} & \textbf{93.50} & 62.21 & \textbf{51.64} & \ul{43.55}\\
\textbf{CAKE} & \textbf{19.83} & \textbf{41.69} & \ul{45.76} & 54.77 & 47.23 & 37.33 & \textbf{25.08} & 21.51 & 22.56 & 21.50 & 0.17 & 24.34 & 69.00 & 90.19 & 40.47 & 23.00 & \textbf{7.00} & \ul{70.00} & \ul{93.00} & \textbf{62.64} & 50.55 & 43.37\\
\textbf{LAVa (Ours)} & 19.35 & \ul{41.52} & 45.53 & 55.49 & 47.25 & \textbf{37.96} & 24.36 & \ul{22.01} & \textbf{23.19} & 21.53 & 0.17 & \textbf{25.38} & \textbf{70.50} & \ul{90.21} & \textbf{41.30} & \textbf{23.75} & \textbf{7.00} & \textbf{70.50} & \textbf{93.50} & 62.04 & \ul{51.57} & \textbf{43.70}\\
\midrule
\multicolumn{23}{c}{\textbf{$\mathbb{B}=1024\mathrm{HL}$}}\\
\textbf{PyramidKV} & 19.73 & 41.62 & 43.51 & 56.75 & \textbf{48.73} & 37.23 & 24.32 & 22.76 & 24.72 & 21.91 & 0.15 & 25.97 & 71.00 & 90.23 & 40.99 & \textbf{23.75} & \ul{6.00} & \textbf{71.00} & \ul{93.00} & 60.66 & 49.31 & 43.66\\
\textbf{SnapKV} & 19.31 & 41.67 & 43.92 & 56.25 & 47.90 & 37.77 & \ul{24.50} & 22.55 & 24.76 & 21.72 & 0.25 & 25.91 & \ul{72.00} & 90.24 & 41.43 & \ul{23.50} & \textbf{6.50} & \textbf{71.00} & \ul{93.00} & 61.59 & \textbf{50.53} & 43.80\\
\textbf{Ada-PyramidKV} & \ul{20.23}& 41.47 & 44.82 & 56.39 & \ul{48.48} & 37.62 & 24.35 & \ul{23.59} & 25.14 & \ul{22.24} & 0.17 & 26.05 & 71.00 & 90.23 & 41.34 & \textbf{23.75} & \ul{6.00} & \textbf{71.00} & \ul{93.00} & 61.01 & 48.65 & 43.82 \\
\textbf{Ada-SnapKV} & 19.99 & \ul{42.07} & \ul{45.86} & \ul{57.16} & 48.29 & 38.01 & 24.40 & \textbf{23.60} & \textbf{25.37} & 22.19 & 0.16 & \ul{26.10} & \ul{72.00} & \ul{90.31} & 41.50 & \ul{23.50} & \textbf{6.50} & \textbf{71.00} & \textbf{93.50} & 61.62 & \ul{50.30} & \ul{44.16}\\
\textbf{CAKE} & 19.29 & 41.67 & 43.72 & 56.38 & 47.70 & \textbf{38.10} & \textbf{24.83} & 22.72 & 24.68 & 21.90 & 0.21 & \textbf{26.25} & \ul{72.00} & 90.24 & \ul{41.59} & \ul{23.50} & \ul{6.00} & \textbf{71.00} & \ul{93.00} & \ul{61.46} & 50.17 & 43.81\\
\textbf{LAVa (Ours)} & \textbf{20.93} & \textbf{42.74} & \textbf{46.43} & \textbf{57.26} & 48.35 & \ul{38.09} & 24.44 & 23.26 & \ul{25.26} & \textbf{22.33} & 0.17 & \textbf{26.25} & \textbf{72.50} & \textbf{90.39} & \textbf{42.05} & \ul{23.50} & \ul{6.00} & \textbf{71.00} & \ul{93.00} & \textbf{61.93} & 50.23 & \textbf{44.30}\\

\bottomrule
\end{tabular}
\end{adjustbox}
\caption{Final comparison Based on Llama3-8B-Instruct among 21 datasets of LongBench. (Note: The best result is highlighted in \textbf{bold}, and the second is in \ul{underline}.)}
\label{our llama3 longbench}
\end{table*}

\paragraph{Results of LAVa in Needle In A Haystack.}
\label{ref:needle}
The results of Needle In A Haystack are shown in Table~\ref{tab:mistral-needle}. The conclusion is consistent with that of LongBench. Our method shows superior overall performance, demonstrating its robust in preserving the model’s retrieval capacity.

\begin{table}[t]
\centering
\begin{adjustbox}{max width=\textwidth}
\begin{tabular}{lcc}
\toprule
 \textbf{Methods} & \textbf{Mistral-7B}  & \textbf{Qwen2.5-7B}  \\
 \midrule
Full Cache & 99.88 & 99.66    \\
\midrule
&  \multicolumn{2}{c}{\textbf{$\mathbb{B}=128\mathrm{HL}$}}\\
\textbf{PyramidKV} & 91.44 & 91.10  \\
\textbf{SnapKV} & 91.25 & 93.28  \\
\textbf{Ada-PyramidKV} & 92.08 & 92.70   \\
\textbf{Ada-SnapKV} & 92.12  & 94.30  \\
\textbf{CAKE} & 92.79  & 94.61 \\
\textbf{LAVa (Ours)} & \textbf{93.35}  & \textbf{95.57}   \\
\midrule
 & \multicolumn{2}{c}{\textbf{$\mathbb{B}=1024\mathrm{HL}$}}\\
\textbf{PyramidKV} & 97.88 & 99.56   \\
\textbf{SnapKV} & 97.95 & 99.48  \\
\textbf{Ada-PyramidKV} & 98.58 & 99.58 \\
\textbf{Ada-SnapKV} & 98.54 & 99.53  \\
\textbf{CAKE} & 98.32  & 99.55\\
\textbf{LAVa (Ours)} & \textbf{98.95}  & \textbf{99.59}  \\
\bottomrule
\end{tabular}
\end{adjustbox}
\caption{Average scores of Mistral-7B-Instruct-v0.2 and Qwen2.5-7B-Instruct in Needle In A HayStack.}
\label{tab:mistral-needle}
\end{table}

 \paragraph{Results of LAVa in Ruler and InfiniteBench.} The results of Ruler and InfiniteBench are shown in Table~\ref{tab:mistral-ruler} and Table~\ref{tab:mistral-infinitebench}. we set the cache budget as 5\%-10\% of the task context length, i.e. 1024 and 10000. We use Mistral-7B-Instruct-v0.2 as the backbone of Ruler. For InfiniteBench, we change the backbone into Mistral-7B-LongPO-128K~\cite{chen2025longpo}, which is fine-tuned based on Mistral-7B-Instruct-v0.2, because the task context length of InfiniteBench is much longer than the original maximum model length 32K. The results reconfirm the effectiveness of LAVa.

\paragraph{Results of Dynamic Budget Allocation.}
\label{detailed ab}
The detailed results of ablation study based on Mistral-7B-Instruct-v0.2 in LongBench are listed in Table~\ref{our ab longbench}. It demonstrates that dynamic budget allocation at both the head and layer levels is essential for strong performance, with a more pronounced performance drop when head-wise allocation is removed under constrained budgets. This is expected, as LAVa’s strength lies in its ability to compare cache entries across heads.


\begin{table*}[htbp]
\centering
\begin{adjustbox}{max width=\textwidth}
\begin{tabular}{lccccccccccccccccccccc|c}
\toprule
\multirow{2}{*}{} & \multicolumn{4}{c}{\textbf{Single-Doc. QA}} & \multicolumn{4}{c}{\textbf{Multi-Doc. QA}} & \multicolumn{4}{c}{\textbf{Summarization}} & \multicolumn{4}{c}{\textbf{Few-shot Learning}} & \multicolumn{3}{c}{\textbf{Synthetic}} & \multicolumn{2}{c}{\textbf{Code}} & \multirow{1}{*}{}\\
\cmidrule(lr){2-5} \cmidrule(lr){6-9} \cmidrule(lr){10-13} \cmidrule(lr){14-17} \cmidrule(lr){18-20} \cmidrule(lr){21-22} 
& \rotatebox{-60}{NrtvQA} & \rotatebox{-60}{Qasper} & \rotatebox{-60}{MF-en} & \rotatebox{-60}{MF-zh} & \rotatebox{-60}{HotpotQA} & \rotatebox{-60}{2WikiMQA} & \rotatebox{-60}{Musique} & \rotatebox{-60}{Dureader} & \rotatebox{-60}{GovReport} & \rotatebox{-60}{QMSum} & \rotatebox{-60}{VCSUM} & \rotatebox{-60}{MultiNews} & \rotatebox{-60}{TREC} & \rotatebox{-60}{TriviaQA} & \rotatebox{-60}{SAMSum} & \rotatebox{-60}{LSHT} &  \rotatebox{-60}{PCount} & \rotatebox{-60}{PR-en} & \rotatebox{-60}{PR-zh} & \rotatebox{-60}{Lcc} & \rotatebox{-60}{RepoBench-P} & \rotatebox{-60}{Avg}\\
\midrule
\textbf{Full Cache} & 26.77 & 32.34 & 49.63 & 48.42 & 43.43 & 27.89 & 18.61 & 30.85 & 32.92 & 24.54 & 15.04 & 27.20 & 71.00 & 86.23 & 43.41 & 39.00 & 2.81 & 86.56 & 89.75 & 55.29 & 52.55 & 45.07\\

\midrule
\multicolumn{23}{c}{\textbf{$\mathbb{B}=128\mathrm{HL}$}}\\
\textbf{LAVa (Ours)} & 19.57 & 21.11 & 44.29 & 33.91 & \textbf{38.29} & 23.59 & \textbf{15.32} & \textbf{18.56} & \textbf{19.33} & \textbf{22.32} & \textbf{11.42} & \textbf{21.07} & \textbf{53.50} & 85.20 & \textbf{40.16} & \textbf{21.75} & 2.88 & \textbf{69.87} & \textbf{74.75} & \textbf{51.94} & \textbf{48.92} & \textbf{36.74} \\
\ \ \ \ \ $-$ layer & 20.32 & \textbf{21.18} & \textbf{45.17} & \textbf{35.00} & 37.37 & \textbf{23.62} & 15.09 & 18.20 & 19.21 & 22.04 & 11.35 & 20.99 & 48.50 & \textbf{85.32} & 39.33 & 20.75 & 3.42 & 67.93 & 73.75 & 51.28 & 47.52 & 36.20\\
\ \ \ \ \ $-$ head & \textbf{20.33} & 20.27 & 44.06 & 
 32.23 & 36.64 & 22.84 & 14.19 & 18.15 & 18.88 & 21.51 & 11.09 & 20.89 & 45.00 & 84.29 & 39.57 & 20.25 & 3.21 & 65.23 & 64.25 & 51.88 & 47.51 & 34.95\\
\midrule
\multicolumn{23}{c}{\textbf{$\mathbb{B}=256\mathrm{HL}$}}\\
\textbf{LAVa (Ours)} & \textbf{22.70} & 24.67 & \textbf{48.62} & \textbf{37.81} & \textbf{39.68} & \textbf{25.96} & \textbf{16.77} & \textbf{20.26} & \textbf{21.92} & 22.48 & \textbf{11.88} & \textbf{22.91} & \textbf{65.00} & 85.24 & 41.28 & \textbf{26.75} & 2.88 & 76.76 & 85.75 & \textbf{54.17} & \textbf{51.77} &\textbf{40.12}\\
\ \ \ \ \ $-$ layer & 21.78 & \textbf{24.74} & 47.82 & 37.47 & 39.06 & 25.53 & 16.21 & 19.94 & 21.86 & \textbf{23.22} & 11.81 & \textbf{22.91} & 62.00 & \textbf{85.37} & \textbf{41.53} & 25.25 & 2.77 & \textbf{78.53} & \textbf{87.67} & 52.78 & 49.85 & 39.77\\
\ \ \ \ \ $-$ head & 21.34 & 22.77 & 47.43 & 35.87 & 37.71 & 25.50 & 15.47 & 19.43 & 21.55 & 23.06 & 12.08 & 22.86 & 58.00 & 84.88 & 41.69 & 22.25 & 3.11 & 74.77 & 84.18 & 53.89 & 51.19 & 38.80\\
\midrule
\multicolumn{23}{c}{\textbf{$\mathbb{B}=512\mathrm{HL}$}}\\
\textbf{LAVa (Ours)} & \textbf{25.01} & 27.84 & \textbf{48.97} & \textbf{42.14} & \textbf{40.95} & \textbf{26.88} & 18.33 & \textbf{21.12} & 23.59 & \textbf{23.59} & 12.28 & \textbf{24.51} & \textbf{68.50} & \textbf{86.34} & \textbf{42.48} & \textbf{33.50}  & 2.90 & 87.23 & \textbf{89.83} & \textbf{55.83} & \textbf{52.85} & \textbf{42.59} \\
\ \ \ \ \ $-$ layer & 24.43 & \textbf{27.98} & 48.72 & 41.00 & 40.23 & 26.17 & \textbf{18.50} & 20.74 & \textbf{24.00} & 23.40 & \textbf{12.68} & 24.20 & 66.50 & 86.04 & 42.26 & 32.75 & 2.84 & \textbf{87.89} & 89.33 & 54.11 & 51.22 & 42.11\\
\ \ \ \ \ $-$ head & 23.59 & 27.70 & 48.61 & 40.61 & 40.22 & 25.79 & 17.87 & 20.68 & 23.91 & 23.39 & 12.38 & 24.28 & 66.50 & 86.09 & 41.95 & 28.50 & 2.97 & 86.88 & 89.17 & 55.73 & 52.53 & 41.82\\
\midrule
\multicolumn{23}{c}{\textbf{$\mathbb{B}=1024\mathrm{HL}$}}\\
\textbf{LAVa (Ours)} & 25.59 & \textbf{31.21} & 48.27 & 43.43 & \textbf{41.92} & 27.38 & \textbf{19.48} & \textbf{23.48} & 26.06 & \textbf{23.86} & \textbf{13.38} & \textbf{26.00} & \textbf{70.00} & 86.22 & 42.43 & \textbf{38.00} & 2.73 & 87.01 & 88.75 & \textbf{57.31} & \textbf{53.28} & \textbf{43.65} \\
\ \ \ \ \ $-$ layer & \textbf{25.76} & 30.38 & \textbf{49.54} & \textbf{43.54} & 41.08 & 27.03 & 18.83 & 22.73 & 25.79 & 23.69 & 13.13 & 25.88 & 69.50 & \textbf{86.30} & \textbf{43.10} & 37.25 & 2.71 & 87.56 & \textbf{89.25} & 55.04 & 51.67 & 43.35\\
\ \ \ \ \ $-$ head & 25.76 & 29.61 & 49.31 & 42.77 & 40.82 & \textbf{27.63} & 18.59 & 22.64 & \textbf{26.29} & 23.77 & 12.70 & 25.82 & 68.00 & 85.82 & 41.77 & 35.00 & 2.63 & \textbf{89.06} & \textbf{89.25} & \textbf{57.31} & 53.22 & 43.26 \\
\bottomrule
\end{tabular}
\end{adjustbox}
\caption{Ablation study based on Mistral-7B-Instruct-v0.2 among 21 datasets of LongBench. (Note: The best result is highlighted in \textbf{bold}. )}
\label{our ab longbench}
\end{table*}

\paragraph{Analysis of Different Layer Allocation.}
\label{detailed diiferent layer allocation}
 To validate the effectiveness of our layer budget allocation, we modify LAVa to incorporate two alternative strategies: \textbf{LAVa-Uniform}, which is equivalent to LAVa (-layer), and \textbf{LAVa-Pyramid}, which retains LAVa's head budget allocation and layer-wise cache eviction but adopts Pyramid for layer allocation. The results in Table \ref{mistral longbench} indicate that our method outperforms these alternatives. Notably, LAVa-Pyramid requires finetuning, whereas the other methods do not. Moreover, LAVa-Pyramid fails to outperform LAVa-Uniform at higher budgets, aligning with the observed comparison between Ada-SnapKV and Ada-Pyramid. This underscores the limitation of heuristic-based designs, which may not always yield optimal results. 


\begin{table}[t]
\centering
\begin{adjustbox}{max width=\textwidth}
\begin{tabular}{lccc}
\toprule
 \textbf{Context Length} & 4K & 8K & 16K   \\ 
\midrule
\textbf{PyramidKV} & 72.55 & 62.02 & 55.42  \\
\textbf{SnapKV} & 70.71 & 61.52 & 55.61   \\
\textbf{Ada-PyramidKV} & 70.80 & 60.83 & 54.95    \\
\textbf{Ada-SnapKV} & 71.14 & 60.31 & 55.05   \\
\textbf{CAKE} & 72.41 & 61.55 & 55.84  \\
\textbf{LAVa (Ours)} & \textbf{75.39}  & \textbf{62.61} & \textbf{56.70}  \\
\bottomrule
\end{tabular}
\end{adjustbox}
\caption{Results of Mistral-7B-Instruct-v0.2 in Ruler.}
\label{tab:mistral-ruler}
\end{table}

\begin{table}[t]
\centering
\begin{adjustbox}{max width=\textwidth}
\begin{tabular}{lccc}
\toprule
 \textbf{Tasks} & En Sum & En MC & En Dia   \\ 
\midrule
\textbf{PyramidKV} & 25.3 & 67.2 & 6.5  \\
\textbf{SnapKV} & 25.1 & 67.2 & 7.0   \\
\textbf{Ada-PyramidKV} & 24.9 & 67.2 & 7.0   \\
\textbf{Ada-SnapKV} & 24.6 & 66.8 & 7.0   \\
\textbf{CAKE} & 24.8 & \textbf{67.8} & 6.6  \\
\textbf{LAVa (Ours)} & \textbf{25.4}  & 66.8 & \textbf{9.5}    \\
\bottomrule
\end{tabular}
\end{adjustbox}
\caption{Results of Mistral-7B-LongPO-128K in InfiniteBench.}
\label{tab:mistral-infinitebench}
\end{table}

\begin{table*}[htbp]
\centering
\begin{adjustbox}{max width=\textwidth}
\begin{tabular}{lccccccccccccccccccccc|c}
\toprule
\multirow{2}{*}{} & \multicolumn{4}{c}{\textbf{Single-Doc. QA}} & \multicolumn{4}{c}{\textbf{Multi-Doc. QA}} & \multicolumn{4}{c}{\textbf{Summarization}} & \multicolumn{4}{c}{\textbf{Few-shot Learning}} & \multicolumn{3}{c}{\textbf{Synthetic}} & \multicolumn{2}{c}{\textbf{Code}} & \multirow{1}{*}{}\\
\cmidrule(lr){2-5} \cmidrule(lr){6-9} \cmidrule(lr){10-13} \cmidrule(lr){14-17} \cmidrule(lr){18-20} \cmidrule(lr){21-22} 
& \rotatebox{-60}{NrtvQA} & \rotatebox{-60}{Qasper} & \rotatebox{-60}{MF-en} & \rotatebox{-60}{MF-zh} & \rotatebox{-60}{HotpotQA} & \rotatebox{-60}{2WikiMQA} & \rotatebox{-60}{Musique} & \rotatebox{-60}{Dureader} & \rotatebox{-60}{GovReport} & \rotatebox{-60}{QMSum} & \rotatebox{-60}{VCSUM} & \rotatebox{-60}{MultiNews} & \rotatebox{-60}{TREC} & \rotatebox{-60}{TriviaQA} & \rotatebox{-60}{SAMSum} & \rotatebox{-60}{LSHT} &  \rotatebox{-60}{PCount} & \rotatebox{-60}{PR-en} & \rotatebox{-60}{PR-zh} & \rotatebox{-60}{Lcc} & \rotatebox{-60}{RepoBench-P} & \rotatebox{-60}{Avg}\\
\midrule
\textbf{Full Cache} & 26.77 & 32.34 & 49.63 & 48.42 & 43.43 & 27.89 & 18.61 & 30.85 & 32.92 & 24.54 & 15.04 & 27.20 & 71.00 & 86.23 & 43.41 & 39.00 & 2.81 & 86.56 & 89.75 & 55.29 & 52.55 & 45.07\\
\midrule
\multicolumn{23}{c}{\textbf{$\mathbb{B}=128\mathrm{HL}$}}\\
\textbf{LAVa-Pyramid} & 19.91 & 20.36 & 44.32 & \textbf{35.06} & 37.68 & 23.58 & \textbf{15.40} & 17.99 & \textbf{19.61} & 22.09 & 10.87 & 21.05 & 52.00 & 84.45 & 40.09 & 20.25 & 2.89 & \textbf{72.32} & \textbf{76.92} & 51.81 & 46.81 & 36.63\\
\textbf{LAVa-Uniform} & \textbf{20.32} & \textbf{21.18} & \textbf{45.17} & 35.00 & 37.37 & \textbf{23.62} & 15.09 & 18.20 & 19.21 & 22.04 & 11.35 & 20.99 & 48.50 & \textbf{85.32} & 39.33 & 20.75 & 3.42 & 67.93 & 73.75 & 51.28 & 47.52 & 36.20\\
\textbf{LAVa (Ours)} & 19.57 & 21.11 & 44.29 & 33.91 & \textbf{38.29} & 23.59 & 15.32 & \textbf{18.56} & 19.33 & \textbf{22.32} & \textbf{11.42} & \textbf{21.07} & \textbf{53.50} & 85.20 & \textbf{40.16} & \textbf{21.75} & 2.88 & 69.87 & 74.75 & \textbf{51.94} & \textbf{48.92} & \textbf{36.74} \\
\midrule
\multicolumn{23}{c}{\textbf{$\mathbb{B}=256\mathrm{HL}$}}\\
\textbf{LAVa-Pyramid} & 21.22 & 23.96 & 47.86 & 37.12 & 38.92 & 24.94 & 16.70 & 19.11 & 21.43 & 22.44 & 11.20 & 22.77 & 62.50 & 85.17 & 41.34 & 23.75 & 3.34 & \textbf{79.07} & 86.58 & 52.25 & 49.70 & 39.40\\
\textbf{LAVa-Uniform} & 21.78 & \textbf{24.74} & 47.82 & 37.47 & 39.06 & 25.53 & 16.21 & 19.94 & 21.86 & \textbf{23.22} & 11.81 & \textbf{22.91} & 62.00 & 85.37 & \textbf{41.53} & 25.25 & 2.77 & 78.53 & \textbf{87.67} & 52.78 & 49.85 & 39.77\\
\textbf{LAVa (Ours)} & \textbf{22.70} & 24.67 & \textbf{48.62} & \textbf{37.81} & \textbf{39.68} & \textbf{25.96} & \textbf{16.77} & \textbf{20.26} & \textbf{21.92} & 22.48 & \textbf{11.88} & \textbf{22.91} & \textbf{65.00} & 85.24 & 41.28 & \textbf{26.75} & 2.88 & 76.76 & 85.75 & \textbf{54.17} & \textbf{51.77} &\textbf{40.12}\\
\midrule
\multicolumn{23}{c}{\textbf{$\mathbb{B}=512\mathrm{HL}$}}\\
\textbf{LAVa-Pyramid} & 24.59 & 27.33 & 48.36 & 40.24 & 39.75 & 26.18 & 18.26 & 20.82 & 23.39 & 23.38 & 12.35 & 24.08 & 67.00 & \textbf{86.66} & \textbf{42.55} & 32.00 & 2.93 & 86.13 & 89.62 & 53.46 & 51.53 & 41.88\\
\textbf{LAVa-Uniform} & 24.43 & \textbf{27.98} & 48.72 & 41.00 & 40.23 & 26.17 & \textbf{18.50} & 20.74 & \textbf{24.00} & 23.40 & \textbf{12.68} & 24.20 & 66.50 & 86.04 & 42.26 & 32.75 & 2.84 & \textbf{87.89} & 89.33 & 54.11 & 51.22 & 42.11\\
\textbf{LAVa (Ours)} & \textbf{25.01} & 27.84 & \textbf{48.97} & \textbf{42.14} & \textbf{40.95} & \textbf{26.88} & 18.33 & \textbf{21.12} & 23.59 & \textbf{23.59} & 12.28 & \textbf{24.51} & \textbf{68.50} & 86.34 & 42.48 & \textbf{33.50}  & 2.90 & 87.23 & \textbf{89.83} & \textbf{55.83} & \textbf{52.85} & \textbf{42.59} \\
\midrule
\multicolumn{23}{c}{\textbf{$\mathbb{B}=1024\mathrm{HL}$}}\\
\textbf{LAVa-Pyramid} & 24.88 & 29.51 & 49.01  & 42.57 & 41.16 & 27.20 & 19.40 & 22.61 & 25.58 & \textbf{24.00} & 13.08 & 25.71 & 68.50 & 86.19 & \textbf{43.19} & 37.00 & 2.67 & \textbf{87.73} & \textbf{90.25} & 54.72 & 51.53 & 43.19\\
\textbf{LAVa-Uniform} & \textbf{25.76} & 30.38 & \textbf{49.54} & \textbf{43.54} & 41.08 & 27.03 & 18.83 & 22.73 & 25.79 & 23.69 & 13.13 & 25.88 & 69.50 & \textbf{86.30} & 43.10 & 37.25 & 2.71 & 87.56 & 89.25 & 55.04 & 51.67 & 43.35\\
\textbf{LAVa (Ours)} & 25.59 & \textbf{31.21} & 48.27 & 43.43 & \textbf{41.92} & \textbf{27.38} & \textbf{19.48} & \textbf{23.48} & \textbf{26.06} & 23.86 & \textbf{13.38} & \textbf{26.00} & \textbf{70.00} & 86.22 & 42.43 & \textbf{38.00} & 2.73 & 87.01 & 88.75 & \textbf{57.31} & \textbf{53.28} & \textbf{43.65} \\
\bottomrule
\end{tabular}
\end{adjustbox}
\caption{Layer allocation comparison based on Mistral-7B-Instruct-v0.2 among 21 datasets of LongBench. (Note: The best result is highlighted in \textbf{bold}. )}
\label{mistral longbench}
\end{table*}

\begin{table*}[htbp]
\centering
\begin{adjustbox}{max width=\textwidth}
\begin{tabular}{lcccccc}
\toprule
 \textbf{Tasks} & Qasper & HotpotQA & Gov Report & TriviaQA & Passage Retrieval ZH & LCC   \\ 
\midrule
\multicolumn{7}{c}{\textbf{Layer 0}}\\
\textbf{AdaKV} & 1.77 & 1.63 & 1.82 & 2.64 & 1.59 & 1.91 \\
\textbf{LAVa} & \textbf{1.61} & \textbf{1.59} & \textbf{1.73} & \textbf{2.61} & \textbf{1.40} & \textbf{1.86}   \\
\midrule
\multicolumn{7}{c}{\textbf{Layer 31}}\\
\textbf{AdaKV} & 134.69 & 133.33 & 107.94 & 121.53 & 93.50 & 149.25 \\
\textbf{LAVa} & \textbf{132.97} & \textbf{130.02} & \textbf{106.06} & \textbf{121.31} & \textbf{90.50} & \textbf{147.16}   \\
\bottomrule
\end{tabular}
\end{adjustbox}
\caption{Results of Layer Attention Output Loss.}
\label{tab:layer attention loss}
\end{table*}

\paragraph{Analysis of Time Complexity.} Our study builds upon the SnapKV framework with a batch size of 1, consistent with prior works like CAKE and AdaKV. We start with the analysis for SnapKV (the most computationally efficient method among baselines) in computation for one layer as a reference.

\begin{itemize}
    \item For layer $l$, SnapKV needs to calculate the layer’s original KV Cache with the time complexity of $O(HN^2d_h)$, ignoring the IO operations. Generally, this is done with FlashAttention, which avoids saving the large attention matrix of size $O(N^2)$. The computation cost in practice is high due to IO operations and recomputation (to avoid saving the attention matrix), but we ignore it for simplicity.
    \item As Flash attention does not save the attention matrix, for calculating the scores to evict KV Cache, SnapKV needs to recompute the attention scores for the recent window of size $w$ in the second pass. The time complexity is $O(HNwd_h)$.
    \item The top-$B_{l,h}$ selection for head-wise cache eviction with a min-heap takes $O(NlogB_{l,h})$, and for $H$ heads, it takes $O(HNlogB_{l,h})$, where $B_{l,h}H=B_l,B_lL=\mathbb{B}$.
\end{itemize}

\noindent To summarize, SnapKV requires:

\begin{itemize}
    \item $O(HN^2d_h)$ for original cache for one layer;
    \item $O(HNwd_h)$ for recomputing the recent attention scores;
    \item $O(HNlogB_{l,h})$ for cache eviction.
\end{itemize}

\noindent In contrast, LAVa requires the computation for one layer as follows:
\begin{itemize}
\item $O(HN^2d_h)$ for the original cache of one layer, same as SnapKV;
\item $O(HNwd_h)$ for recomputing the recent attention scores, same as SnapKV;
\item $O(HNd_h)$ for computing the value norms for each token;
\item $O(HNlogB_l)$ for layer-wise cache eviction because the eviction of LAVa is operated in all cache of one layer.
\end{itemize}

\noindent For one layer $l$, the difference of time complexity between LAVa and SnapKV is $O(HN(d_h+logH)$. In a long context, $N$ is very large, and thus $O(HN(d_h+logH)$ is much smaller than the dominant factor $O(HN^2d_h)$. Based on the setting of Mistral-7B-Instruct-v0.2, we have $d_h=128$ and $H=32$, \textbf{the extra computation of LAVa compared to SnapKV is $HN(d_h+logH)$ divided by $HN^2d_h$, which is approximately 0.01\% when $N=10,000$}. The computation time increases with the increase of the number of layers and batch size for both SnapKV and LAVa, but the ratio of the extra computation time for LAVa is still 0.01\%. A similar analysis can be achieved to see that all the other methods have similar latency, aligning with the latency results in Figure~\ref{peak memory}. 

\paragraph{Analysis of Memory Usage.} We analyze the difference between SnapKV and LAVa/CAKE, which are dynamic layer budget methods.
\begin{itemize}
\item For SnapKV, the cache size increases from $O(HKd_h)$ in the first layer to the last layer, where it reaches the peak of $O(LHB_{l,h}d_h)$. The memory peaks when the latest (full) layer cache $O(HNd_h)$ is not pruned, and the current retained cache reaches the size of $O(LHB_{l,h}d_h)$. In sum, the peak memory is $O(HNd_h + LHB_{l,h}d_h)$.

\item For LAVa and CAKE, the cache size is always $O(LHB_{l,h}d_h)$ from the first layer to the last layer, yet it is distributed among prefilled layers. The memory peak, however, is similar to SnapKV, which is $O(HNd_h + LHB_{l,h}d_h)$, except that for LAVa/CAKE, we need to store the layer scores. As we save only the top scores for each layer, the size for scores is $O(LHB_{l,h})$. Given that the total cache size is $O(LHB_{l,h}d_h)$, it is sufficient to just keep a total of $LHK$ scores for comparison. Again, the extra factor is dominated by $O(HNd_h + LHB_{l,h}d_h)$. \textbf{The extra memory usage of LAVa is 0.6\% of SnapKV peak memory} when $L=H=32, B_{l,h}=1024, d_h=128$, and $N=10,000$. This is small, but not as negligible as in time complexity, consistent with Figure~\ref{peak memory}. However, dynamic layer budget is important for tasks like summarization or code generation, as shown in Figure~\ref{figure:generation_extraction}.
\end{itemize}

\paragraph{Analysis of Layer Attention Output Loss.} To validate the effectiveness of LAVa in minimizing layer attention output loss, we compare LAVa with AdaKV, which also aims to minimize layer attention output loss and its scoring function is the same with SnapKV. We set the cache budget as 128 to make the difference clear and calculate the loss in the first and the last layer. The backbone is Mistral-7B-Instruct-v0.2. The results in Table~\ref{tab:layer attention loss} are consistent with the evaluation of other benchmarks, proving that the upper bound of LAVa is tighter compared to that of AdaKV.

\end{document}